\documentclass[preprint,review,12pt]{elsarticle}

\usepackage{epsfig}
\usepackage{amssymb}
\usepackage{amsmath}
\usepackage{amsthm}
\usepackage[colorlinks]{hyperref}       
\usepackage{url}            
\usepackage{booktabs}       
\usepackage{amsfonts}       
\usepackage{nicefrac}       
\usepackage{microtype}      
\usepackage{xcolor}         
\usepackage{enumitem}
\usepackage{graphicx}
\usepackage{caption}
\usepackage[table]{xcolor}
\usepackage{colortbl}
\usepackage{subcaption}
\usepackage{wrapfig}        
\usepackage{tabularx}   
\usepackage{booktabs}   
\usepackage{ragged2e}   
\definecolor{pastelgreen}{rgb}{0.75, 0.95, 0.75}
\definecolor{pastelblue}{rgb}{0.75, 0.85, 0.95}
\definecolor{amber}{rgb}{1.0, 0.68, 0.0}
\definecolor{lavenderblue}{rgb}{0.8, 0.8, 1.0}
\definecolor{darkbrown}{rgb}{0.4, 0.26, 0.13}
\newcommand{\formattedparagraph}[1]{\noindent \textbf{#1}}
\newcommand{\revisedone}[1]{\textcolor{black}{#1}}
\newcommand{\secondrevisedone}[1]{\textcolor{black}{#1}}

\newcommand\blfootnote[1]{
  \begingroup
  \renewcommand\thefootnote{} 
  \footnote{#1}
  \endgroup
}

\usepackage{lineno}

\journal{Computer Vision and Image Understanding}

\begin{document}

\begin{frontmatter}



\title{Time-Archival Camera Virtualization for Sports and Visual Performances}

%

\author[label1,label2,label5]{Yunxiao Zhang} 
\author[label4,label5]{William Stone} 
\author[label1,label2,label3,label4,label5]{Suryansh Kumar{$^{\dagger}$\blfootnote{$^{\dagger}$Corresponding Author}}} 

\affiliation[label1]{organization={Visual and Spatial AI Lab, Visual Computing \& Computational Media (VCCM) Section}}
\affiliation[label2]{organization={College of Performance, Visualization, and Fine Arts (PVFA)}}
\affiliation[label3]{organization={Department of Electrical and Computer Engineering (ECEN)}}
\affiliation[label4]{organization={Department of Computer Science and Engineering (CSCE)}}
\affiliation[label5]{organization={Texas A\&M University}, city={College Station}, state={Texas}, country={USA}}

\begin{abstract}
Camera virtualization—an emerging solution to novel view synthesis—holds transformative potential for visual entertainment, live performances, and sports broadcasting by enabling the generation of photorealistic images from novel viewpoints using images from a limited set of calibrated multiple static physical cameras. Despite recent advances, achieving spatially and temporally coherent and photorealistic rendering of dynamic scenes \revisedone{with efficient time-archival capabilities}, particularly in fast-paced sports and stage performances, remains challenging for existing approaches. Recent methods based on 3D Gaussian Splatting (3DGS) for dynamic scenes \secondrevisedone{could} offer real-time \revisedone{view-synthesis results}. Yet, they are hindered by their dependence on accurate 3D point clouds from the structure-from-motion method and their inability to handle \revisedone{large, non-rigid, rapid motions of different subjects (e.g., flips, jumps, articulations, sudden player-to-player transitions). Moreover, independent motions of multiple subjects can break the Gaussian-tracking assumptions commonly used in 4DGS, ST-GS, and other dynamic splatting variants}. \revisedone{This paper advocates reconsidering a neural volume rendering formulation for camera virtualization and efficient time-archival capabilities, making it useful for sports broadcasting and related applications}. By modeling a dynamic scene as rigid transformations across multiple synchronized camera views at a given time, our method performs neural representation learning, providing enhanced visual rendering quality at test time. A key contribution of our approach is its support for \textbf{time-archival}, i.e., users can revisit any past temporal instance of a dynamic scene and can perform novel view synthesis, enabling retrospective rendering for replay, analysis, and archival of live events—a functionality absent in existing neural rendering approaches and novel view synthesis methods for dynamic scenes. \secondrevisedone{While, in principle, dynamic 3DGS approaches can also perform time-archival, however, it will require either a multi-view structure-from-motion (SfM) point cloud to be stored at every time step or some form of additional multi-body temporal modeling constraint---both of which are complex, computationally expensive, and could be memory-intensive.}
\secondrevisedone{We argue that a dynamic scene observed under a well-constrained synchronized multiview setup---typical in sports and visual performance scenarios, is already strongly constrained by geometry, and we may not need a temporally coupled constraint or 3d point cloud initialization.}
Extensive experiment and ablations on established benchmarks and our newly proposed dynamic scene datasets demonstrate that our method surpasses 4DGS-based baselines in rendered image quality and \revisedone{other performance metric for time-archival view-synthesis for a dynamic scene}, thus setting a new standard for virtual camera systems in dynamic visual media. Furthermore, our approach could be an encouraging step towards compactly modeling the plenoptic function, allowing for time-archival of a long video sequence.
\end{abstract}
\begin{graphicalabstract}  
 \begin{figure*}[h]
    \centering
    \includegraphics[width=0.9\textwidth]{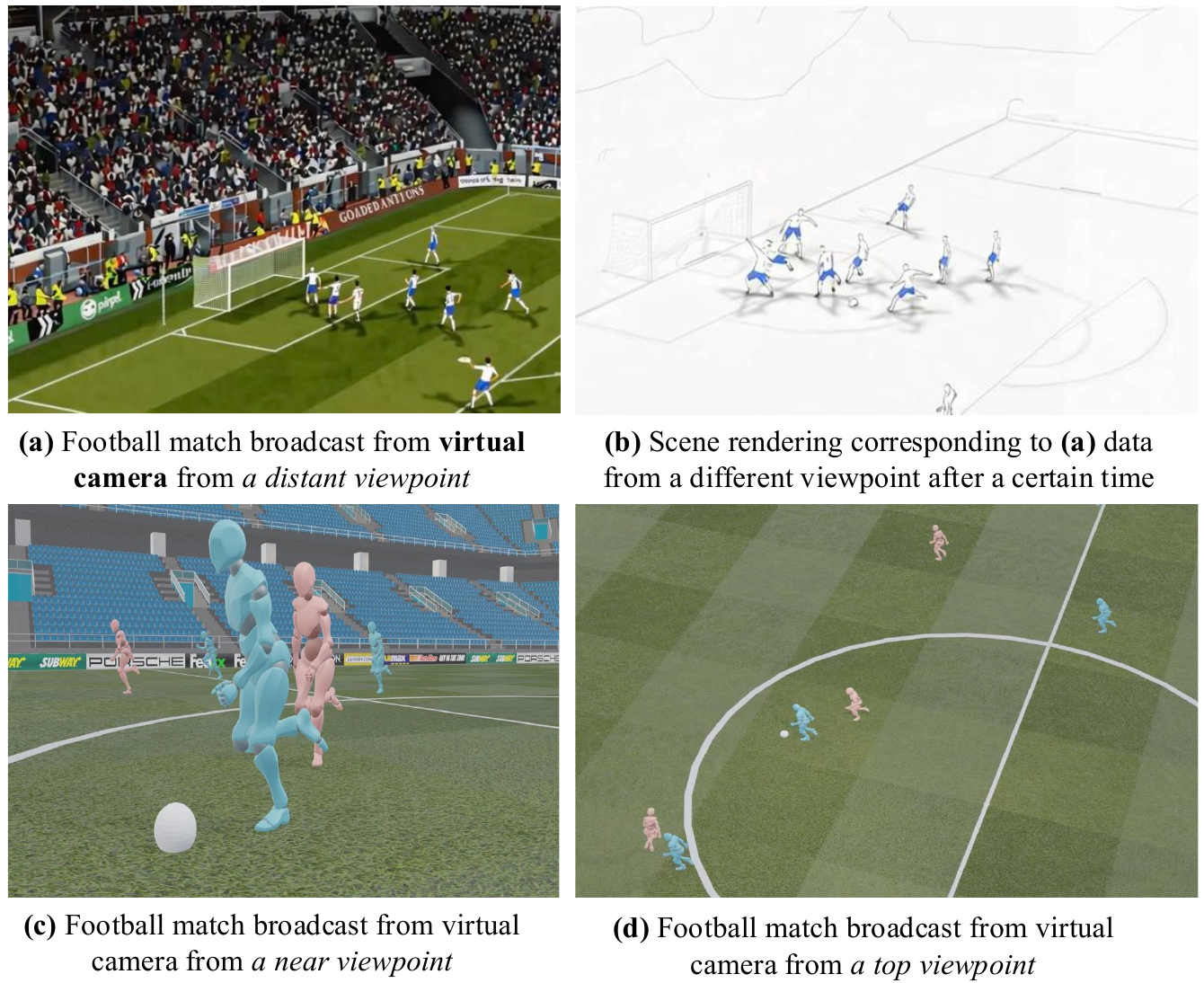}
    \caption{\small \label{fig:graphical_abstract} (a)-(d) Camera virtualization for football sports showing the image rendering from camera placed at different distances from the subject(s), i.e., (a) far-distance viewpoint, (c) near-distance viewpoint (d) top viewpoint.}
  \end{figure*}

Camera virtualization enables photorealistic rendering of dynamic scenes from novel viewpoints using limited static camera setups, significantly benefiting visual entertainment, live performances, and sports broadcasting. Current dynamic-scene rendering methods, particularly those based on 3D Gaussian Splatting (3DGS), enable real-time image synthesis but often depend on high-quality initial 3D point clouds, making them unsuitable for \textbf{time-archival} for \secondrevisedone{our targeted applications}. To overcome these limitations, we revisit the neural implicit scene representation and propose a neural volume rendering framework grounded in multiview projective geometry. \secondrevisedone{We argue that a dynamic scene observed under a well-constrained synchronized multiview setup, typical in sports and visual performance scenarios, is already strongly constrained by geometry, and we may not need a temporally coupled constraint or 3d point cloud initialization}. Our method exploits such a default design choice and represents dynamic scenes using synchronized neural representations across multiple camera views, naturally supporting efficient temporal archival and retrospective novel-view synthesis. Experiments on standard benchmarks and newly introduced dynamic-scene datasets demonstrate that our method achieves superior rendering quality compared to state-of-the-art 4DGS approaches, establishing a new benchmark for camera virtualization in dynamic visual media. This work also advances compact plenoptic scene modeling, enabling comprehensive archival and replay of dynamic events.
\end{graphicalabstract}

\begin{highlights}
\item Implicit neural scene representation learning to compactly store temporal instances, allowing users to ``rewind" and synthesize novel views of past and current moments.

\item An approach that represents a dynamic scene using simple neural network model over discrete time step with excellent image-rendering quality compared to the state-of-the-art approaches.

\item \secondrevisedone{While alternative representations (such as 3DGS) could, in principle, support similar functionality, we show that, under the deliberate design choice of a synchronized multiview camera setup typical of sports and visual performance capture, the problem is geometrically well-constrained. To this end, the proposed approach offers a better alternative for time-archival and retrospective novel-view synthesis.}

\item Presents a new synthetic dataset targeted at visual performance and sports applications and benchmarks the related state-of-the-art methods on this dataset.
\end{highlights}


\end{frontmatter}


\section{Introduction}\label{sec:Introduction}
Sporting events and visual performances are inherently dynamic in nature. More importantly, the dynamic subjects are of significant interest here compared to other parts of the scene. Traditional methods of capturing such events typically involve physical multi-view camera setups constrained by fixed viewpoints, limited spatial coverage, and logistical complexities. These constraints restrict the freedom to fully exploit the rich visual dynamics intrinsic to sports and visual performance such as dance events, and others. In such applications, we can greatly enhance viewer engagement by allowing the user to watch the event from novel viewpoints. Furthermore, such an application must support time-archival, i.e., the user can revisit the scene in the past and can observe its dynamics from a novel viewpoint. We term this problem here as ``camera virtualization''. Although this problem closely aligns with dynamic scene novel view synthesis, where the goal is to render a photorealistic image of dynamic subject(s) from a novel camera viewpoint at test time, camera virtualization \textbf{must allow for time-archival}. 
\begin{wrapfigure}{R}{0.45\textwidth}
\centering
\includegraphics[width=0.45\textwidth]{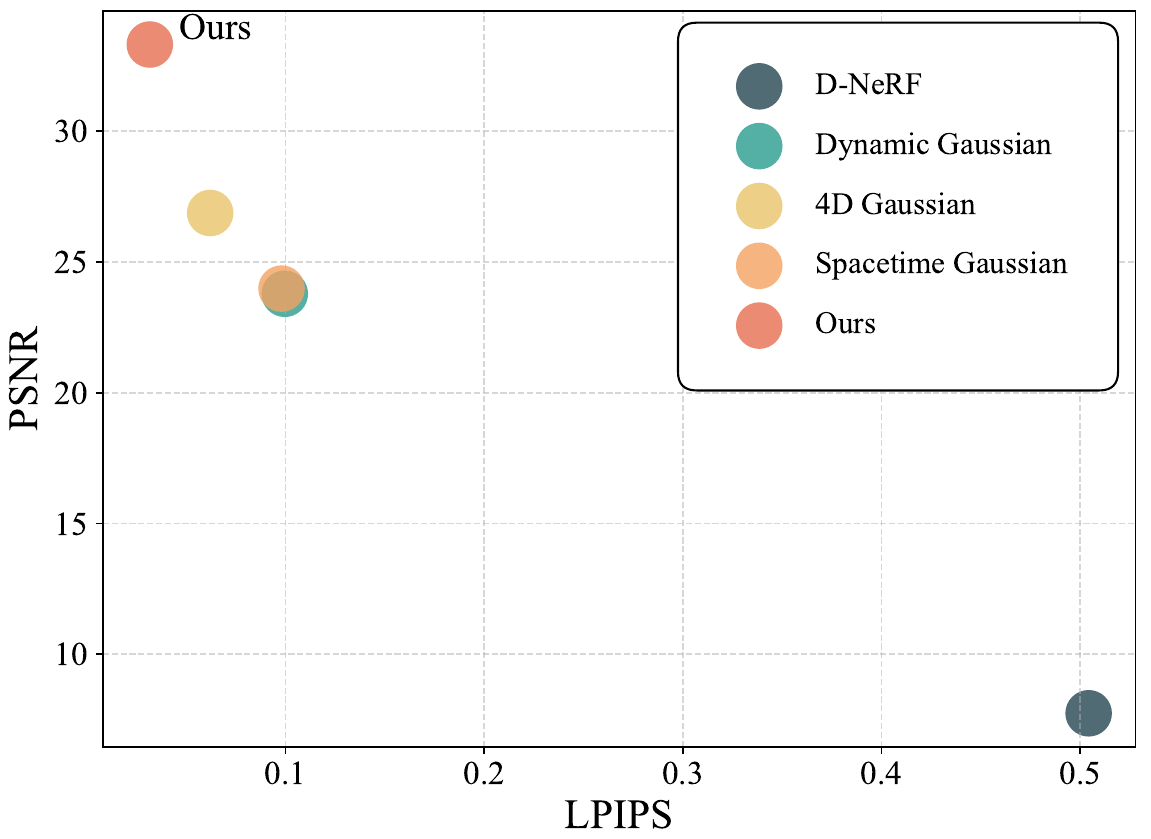}
\caption{\small \label{fig:lpips_psnr_comparison}Quantitative comparison of the image rendering quality on the proposed dynamic scene dataset with state.}
\end{wrapfigure}
By synthesizing novel visual perspectives of a dynamic scene from limited sets of physical cameras, camera virtualization enables unprecedented flexibility in visual content generation. To this end, recent advances in neural rendering techniques such as Neural Radiance Fields (NeRF) \cite{mildenhall2021nerf} and 3D Gaussian splatting (GS) \cite{kerbl20233d} and its extensions to dynamic scenes \cite{wu20244d, li2024spacetime} could provide a promising solution. Yet, significant technical challenges remain unresolved despite excellent progress, particularly regarding dynamic scene rendering quality, spatial-temporal coherence, and time-archival \revisedone{(compact and memory efficient representation of the dynamic scene for replay from novel viewpoints)}. This paper investigates these challenges and introduces an approach to camera virtualization, explicitly focused on visual performance and sports applications.

Since time-archival is one of the prime motivation of this paper, we advocate that despite the 3D-GS \cite{kerbl20233d} and its extension to the dynamic scene such as 4D-GS \cite{wu20244d} and Spacetime-GS \cite{li2024spacetime} being encouraging in providing real-time rendering performance, novel view synthesis solution specifically for dynamic scenes should not wholly rely on GS based approaches. Instead, exploiting physical camera setup configuration (such as synchronization, view angles, etc.), multiview rigidity constraint, and benefits of neural scene representation learning in image rendering must not be ignored. Not that the rendering speed will not suffer a bit due to such a take, yet the reason for such research exploration lies on the fact the performance of 3DGS based approaches observed to rely heavily on the accuracy of initial 3D points recovered from structure from motion pipeline such as COLMAP \cite{schonberger2016structure} \revisedone{or active sensing modality such as LiDAR \cite{shen2025lidar} or RGB-D \cite{wei2024gsfusion}}. On the contrary, popular NeRF-based methods \cite{mildenhall2021nerf, muller2022instant} do not rely on explicit 3D points for image-based rendering at the same time could help memorize the scene representation compactly over time while also providing accurate image rendering quality. 
\revisedone{Let's understand how 3d point cloud dependent representation such as 3DGS and related approaches leads to several practical challenges in dynamic scene time-archival for applications such as sports.}

\begin{itemize}
    \item \revisedone{Storage Efficiency: A 3DGS model if used for time-archival in our scenarios will typically requires 1-5 million Gaussians per scene, with each Gaussian storing position, covariance, opacity, and dozens of spherical harmonics coefficients. This results in 200-300MB per time step. Thus, for long sequences, per-time-step 3DGS requires tens of gigabytes for storage (e.g., 100 frames: 20-30 GB). In contrast, our proposition for using implicit per-frame radiance field needs ~12.7M parameters (approx. 25-50MB). Therefore, the memory storage usage per time step is 10–20x smaller, allowing time-archival over long sequences. This makes the proposed  formulation  significantly more scalable while keeping the memory footprint predictable and independent of scene complexity.} 
    
    \item \revisedone{For a dynamic scene, if 3DGS is applied for time-archival, it would, in principle, require either a multi-view structure-from-motion (SfM) point cloud at every time step or some form of additional multi-body temporal modeling constraint---both of which are complex, computationally expensive, and, frankly, memory-intensive.}
    
    \item \revisedone{Exact vs. approximated state retrieval: 3DGS extensions to dynamic scene such as 4DGS and others typically rely on tracking or deforming a shared set of Gaussians. Even with per–time-step Gaussian storage, parameter reuse and propagation can lead to drift or approximation errors, especially at time steps distant from keyframes. Our method on the other hand stores a complete radiance function $\mathbf{F}_t$ independently for each time step, enabling exact, drift-free reconstruction for each time instance.}

    \item \revisedone{Independent per-frame optimization avoids compounding errors, i.e., Dynamic 3DGS updates Gaussians sequentially, meaning errors may accumulate across time. On the contrary, our method avoids this entirely by reinitializing the optimization per timestep. Yet, our method stores dense radiance fields that are temporally independent—an advantage in multi-subject, extreme-motion sports environments.}
    
\end{itemize}

This brings us to the point of justifying the above notions and why it makes sense to revisit neural rendering approaches for sports and visual performance applications, which is the paper's primary focus. Firstly, in such applications, we are often provided with a multiview synchronized static camera setup. Therefore, any common dynamic subject(s) between viewpoints at a given time will be rigidly related. This allows us to model a dynamic scene at a given time using an implicit neural representation without the requirement of 3D points as input. Such an approach is easy, reliable, and favorably fast with inherent time-archival capability, i.e., time indexing of neural implicit scene representation. 

\begin{figure*}[t]
\centering
\includegraphics[width=1.0\textwidth]{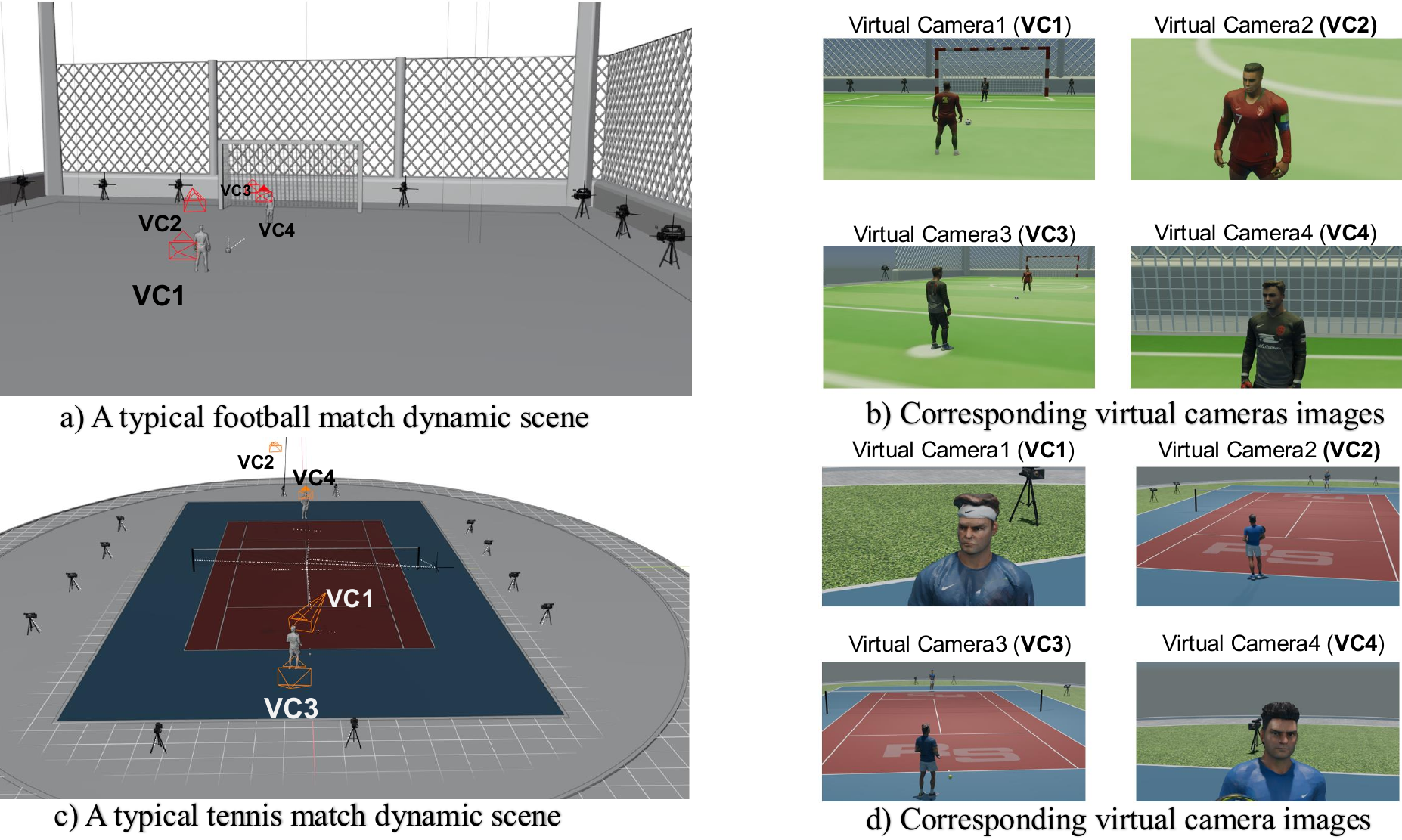}
\caption{Overall setup for our application. A typical multiview synchronized camera setup installed in a sports scene for broadcasting (shown in black). The virtual cameras to inspect the scene are shown in red color for time-archival or broadcast from users interest view points. \textbf{a)-b)Top-row}: A typical football sports scene and image-rendering from virtual cameras. \textbf{c)-d)Bottom-row}:  A typical tennis sports match scene and respective image-rendering from virtual camera view points.}\label{fig:intro-fig}
\end{figure*}

By unifying geometric ideas with neural rendering, our approach can transform applications like sports broadcasting, enabling users to analyze scenes from optimal angles, or theatrical archives preserving performances as 4D experiences. This work sets a new standard for dynamic scene virtualization with inherent time-archival capability (see Figure \ref{fig:lpips_psnr_comparison} for quantitative result comparison with current state-of-the-art methods). So, conceptually, our approach can be thought of as a way to \revisedone{compactly} model plenoptic function $\Phi$  \cite{bergen1991plenoptic, lippman1980movie}
\begin{equation}\label{eq:plenoptic_func}
    \Phi(\mathbf{x}, {\Omega}_{\theta}, {\Omega}_{\phi}, \lambda, t), ~~\textrm{where,} ~\mathbf{x} = (x, y, z)^{T} \in \mathbb{R}^{3},  ~{\Omega}_{\theta} \in [0, \pi],  ~{\Omega}_{\phi} \in [0, 2\pi).
\end{equation}
Here, $\lambda$ is the wavelength which we assume as a constant, and $t$ for time. Contrary to  \cite{li2020crowdsampling, liu2020learning}, we are interested in modeling the notion of $\Phi$ tailored for a dynamic scene, where subject(s) moves through time. Here, $\mathbf{x}$ is a ray position, $(\Omega_\theta, \Omega_\phi)$ is the ray direction. So, if we could model $\Phi$, we could visually reconstruct every possible view, at every moment, from every position. Fig. \ref{fig:intro-fig} shows a couple of examples 
use of our approach for sports. In this paper, we make the following contributions:

\begin{enumerate}[leftmargin=*, noitemsep]
    \item The paper proposes the notion and solution to camera virtualization with time-archival capabilities. By revisiting the implicit neural scene representation learning to compactly store temporal instances of a dynamic scene, this paper allows users to ``rewind'' and synthesize novel views of past and current moments. Consequently, enabling a key step in compact modeling of the plenoptic function \cite{bergen1991plenoptic, lippman1980movie}.
    \item An approach to represent dynamic scene using simple neural network model over discrete time step with better image-rendering quality compared to the state-of-the-art approaches such as \cite{wu20244d, li2024spacetime}, is proposed. Furthermore, the approach does not need explicit 3d points to model dynamic subjects' \revisedone{large, non-rigid, rapid motions across subjects in synchronized views (e.g., flips, jumps, articulations, sudden player-to-player transitions)}.

    \item \secondrevisedone{While alternative representations (such as 3DGS) could, in principle, support similar functionality, we show that, under the deliberate design choice of a synchronized multiview camera setup typical of sports and visual performance capture, the problem is geometrically well-constrained. To this end, the proposed approach offers a better alternative for time-archival and retrospective novel-view synthesis.}
    
    \item The paper introduces a new synthetic dataset targeted at visual performance and sports applications and benchmarks the related state-of-the-art methods on this dataset.
\end{enumerate}

\section{Related Work}\label{sec:related-works}

Image-based rendering (IBR), now popularly re-branded as novel view synthesis, has long been a central problem in computer graphics and computer vision. This research area has a rich body of classical literature \cite{seitz1996view, buehler2001unstructured}. For this paper, we limit our discussion to recent papers that are most relevant to our work. For broad literature survey, we refer the readers to \cite{xiao2025neural, 10.1111:cgf.14505, chen2024survey}.

\formattedparagraph{\textit{(i)} Neural Scene Representations.} Neural representations for a 3D scene have significantly advanced novel view synthesis by learning radiance fields from sparse multi-view observations. Early methods map learnable features to volumes \cite{lombardi2019neural}, texture maps \cite{chen2020neural, thies2019deferred}, or point clouds \cite{aliev2020neural}. Not long ago, NeRF \cite{mildenhall2021nerf} introduced differentiable volume rendering via MLPs without the need for explicit scene geometry. 
This marks NeRF \cite{mildenhall2020nerf} as a foundational milestone in neural scene representation. Extending NeRF, numerous works were proposed to improve computational efficiency \cite{muller2022instant} and expressiveness \cite{jain2023enhanced, riegler2021stable}: some optimize ray sampling to reduce point queries \cite{attal2023hyperreel, neff2021donerf}, while others adopt light field-based representations \cite{attal2022learning, feng2021signet, li2021neulf, suhail2022light, wang2022r2l, kaya2021uncalibrated, kaya2022neural} and multi-scale representation \cite{barron2021mip, barron2022mip, jain2022robustifying, jain2024learning}. Another direction extending NeRF introduced explicit, localized encoding for time efficiency \cite{barron2023zip, chen2022tensorf, chen2021multiresolution, fridovich2022plenoxels, hu2023tri, liu2020neural, muller2022instant, reiser2021kilonerf, sun2022direct, takikawa2021neural, xu2022point, yu2021plenoctrees}, including Instant-NGP \cite{muller2022instant} with multiresolution hash grids and TensoRF \cite{chen2022tensorf} using tensor decomposition for compactness.

\formattedparagraph{\textit{(ii)} 3D Gaussian Splatting.} A  recent alternative to neural implicit models is 3D Gaussian Splatting (3DGS) \cite{kerbl20233d}. It represents scenes via anisotropic 3D Gaussians defined by position, covariance, opacity, and view-dependent color. These are rendered directly through a forward rasterization pipeline, achieving real-time synthesis with competitive image quality. Yet, this real-time rendering speed comes with trade-offs, i.e., 3DGS lacks a learnable memory structure and temporal indexing, making it unsuitable for time-archival tasks where revisiting previous moments is critical. Moreover, it heavily depends on accurate 3D point initialization from SfM \cite{schonberger2016structure}, which has well-known limitations dealing with complex dynamic scenes \cite{hartley2003multiple}.

\formattedparagraph{\textit{(iii)} Novel View Synthesis for Dynamic Scenes.} Extending static representations to dynamic settings has attracted considerable interest. Broxton et al. \cite{broxton2020immersive} introduce multi-sphere images bootstrapped into layered meshes. \cite{bansal20204d} decompose static and dynamic regions for screen-space video manipulation. Other methods incorporate view, time, or lighting parameters into learned 2D encodings \cite{bemana2020x}, or leverage multi-sphere models for depth and occlusion resolution in 360-degree video \cite{attal2020}. Lin et al. \cite{lin2021deep, lin2023view} introduce 3D mask volumes to enforce temporal coherence. Neural Volumes \cite{lombardi2021mixture} and its extensions decode 3D fields using encoder-decoder architectures.

Dynamic NeRF-style models extend implicit neural representations to time-varying scenes \cite{li2022neural, wang2022fourier, li2022streaming, song2023nerfplayer, peng2023representing, wang2023neural, shao2023tensor4d, fridovich2023k, cao2023hexplane, attal2023hyperreel, Wang2023ICCV, icsik2023humanrf, lin2023im4d, wang2024masked, Kim2024Sync}. DyNeRF \cite{li2022neural} uses time-conditioned latent codes, StreamRF \cite{li2022streaming} models inter-frame differences, and NeRFPlayer \cite{song2023nerfplayer} decomposes fields into static and deformable components. Tensor4D \cite{shao2023tensor4d}, HexPlane \cite{cao2023hexplane}, and others factorize space-time volumes for efficient modeling. Complementary to these are monocular dynamic view synthesis methods \cite{li2021neural, pumarola2021d, gao2021dynamic, park2021nerfies, tretschk2021non, wang2021neural, xian2021space, du2021neural, park2021hypernerf, fang2022fast, gao2022monocular, liu2023robust, Li2023CVPR} or multi-view neural view synthesis, which rely on priors over motion \cite{jain2024learning}, scene flow \cite{li2021neural}, or depth \cite{jain2023enhanced, liu2023single} to compensate for sparse camera views while maintaining image rendering results. In contrast, our framework exploits synchronized multi-view imagery to better constrain dynamic scene for novel view synthesis.

More recently, 3DGS has been extended to dynamic settings \cite{4k4d2023, luiten2024dynamic, yang2023real, yang2023deformable, wu20244d, xie2023physgaussian, kratimenos2024dynmf, huang2023scgs, lin2023gaussianflow}. 4K4D \cite{4k4d2023} combines 4D point clouds with K-Planes \cite{fridovich2023k} and depth peeling. \cite{luiten2024dynamic} and \cite{yang2023real} employ 4D Gaussians with temporally discrete keyframes and linear motion. Despite their favorable performance, these methods suffer from flickering, temporal inconsistency \cite{kumar2016multi}, limited expressiveness \cite{kumar2016multi, kumar2017spatio}, despite lighter model. Furthermore, most, if not all, require SfM-based 3d point initialization, which is error-prone in dynamic scenes. \revisedone{Not long ago, MCMC-GS \cite{kheradmand20243d} and structure from motion 3d point-free approaches to 3DGS \cite{foroutan2024evaluating} are proposed that could work with inaccurate random or sparse  point cloud initialization. Yet, these approaches becomes unstable when motion lacks smooth temporal continuity, leading to unreliable propagation of Gaussians across frames. For instance, precise per-frame 3D point recovery and tracking become significantly more error-prone in dynamic scene \cite{kumar2017monocular, kumar2019superpixel} such as sports, where limb articulation, occlusion, and pose discontinuity are common. Therefore, after testing on several dynamic sports scene with multi subject sports motions, where subjects undergo high acceleration and non-rigid deformation using different approaches 3D priors \cite{kumar2022organic, kumar2020non}, \emph{requirement of accurate explicit 3D point module becomes an indispensable choice for 3DGS based pipeline to maintain excellent novel view synthesis results}. This raises the primary concern that maintaining image-synthesis quality requires storing accurate 3D points per instance, which results in either a high memory footprint or degraded performance when 3DGS-based approaches are used for our dynamic scene time-archival objective. A recent work QUEEN \cite{girish2024queen} compresses 3DGS dynamic scenes yet relies on Gaussian identity consistency, tracked primitives, canonical motion fields, which  degrades under large motions---a common case in sports applications.}

Our approach departs from prior work in several ways. First, we propose a fully implicit, temporally indexed MLP representation learned from synchronized multi-view inputs, completely eliminating the reliance on SfM-derived 3D points. Second, our method supports compact time-archival for reliable view synthesis, i.e., novel view synthesis at past moments, allowing efficient modeling plenoptic function for dynamic subjects moving through space-time contrary to the static scene papers \cite{li2020crowdsampling, liu2020learning}. This capability is essential for sports replay and visual performance applications and is absent in current 3DGS and NeRF-based approaches. \revisedone{Additionally, unlike \cite{huang2025echoes}, our approach do not need pre-reconstructed accurate T-pose priors for each player for novel view synthesis of a sports scene.} 

\revisedone{Lastly, we note that because each radiance field is modeled independently, training our approach is trivially parallelizable across GPUs, enabling a distributed computing framework that provides exact, drift-free reconstruction for every moment. While 3DGS approaches bring many benefits---without taking anything away from their tremendous value to novel view synthesis research---in the context of sports applications for time-archival, we observe that per-time-step optimization combined with time-archival management is more practical with the proposed implicit formulation than with sequential dynamic 3DGS pipelines, which rely on temporal dependencies that hinder full parallelization and complicate time-archival in dynamic settings.
}

\section{Methodology}\label{sec:methodology}
Central to our work is the goal to support the time-archival of a dynamic event, which leads to this notion of modeling the plenoptic function $\Phi(\mathbf{x}, {\Omega}_{\theta}, {\Omega}_{\phi}, \lambda, t)$ (refer to Eq.\eqref{eq:plenoptic_func}). In our formulation, we assume constant $\lambda$ to focus on RGB color synthesis that maintains a temporally indexed functional scene representation $F_t$ for each time instance. This design allows for time-archival of dynamic scenes, enabling novel-view synthesis for any prior or current moment. Before, we delve into formulation, let's understand our scene acquisition setup.

\formattedparagraph{\textit{(i)} Multi-View Acquisition Setup.}
We begin with $N$ synchronized multi-view static cameras arranged strategically around the scene to ensure comprehensive spatial coverage and minimize occlusion, which is a typical setup for sports and visual performance scene. At each discrete time instance $t$, the setup captures $N$ synchronized multi-view images:
\begin{equation}\label{eq:1}
\mathcal{I}_t = \{I_t^{(1)}, I_t^{(2)}, \dots, I_t^{(N)}\},
\end{equation}
where $I_t^{(i)} \in \mathbb{R}^{H \times W \times 3}$ represents the RGB image captured by the $i^{\text{th}}$ camera at time $t$, with image dimensions $H \times W$.

Additionally, each camera's intrinsic calibration matrix $\mathbf{K}_i \in \mathbb{R}^{3 \times 3}$ and extrinsic parameters (rotation $\mathbf{R}_i \in \mathrm{SO}(3)$ and translation $\mathbf{t}_i \in \mathbb{R}^3$) are known or estimated using classical SfM method \cite{schonberger2016structure}.

\formattedparagraph{\textit{(ii)} Neural Implicit Representation.}
We put-forward a neural implicit representation approach to model the dynamic 3D scene. This representation models the scene implicitly via a multilayer perceptron (MLP) augmented with a spatial hashing-based encoding to facilitate efficient scene representation and real-time rendering. Similar to \cite{muller2022instant}, we  use an input encoding mechanism based on a multi-resolution hash grid. These mappings are efficiently implemented via a hash function that indexes a fixed-size table, allowing a compact yet expressive encoding of spatial detail. The features of each resolution level are concatenated and serve as the input to a lightweight multilayer perceptron (MLP). The MLP itself is intentionally kept shallow and narrow, usually consisting of two to three hidden layers with modest width, making it highly efficient to train and evaluate (refer to \ref{Asubsec:model_training} for technical details on software implementation). Formally, the scene at each time instance $t$ is encoded as a continuous volumetric function:
\begin{equation}
F_t: (\mathbf{x}, \mathbf{d}) \rightarrow (\mathbf{c}, \sigma),
\end{equation}
where $\mathbf{x} \in \mathbb{R}^3$ denotes a 3D spatial location, $\mathbf{d} \in \mathbb{S}^2$ is the viewing direction, $\mathbf{c} \in \mathbb{R}^3$ is the predicted RGB color at point $\mathbf{x}$ viewed from direction $\mathbf{d}$, and $\sigma \in \mathbb{R}^+$ represents the volume density.

Unlike NeRF method and its extensions, which construct a single static scene representation, our approach maintains a separate set of parameters $\Theta_t$ for each time step $t$. This enables temporal indexing, thus preserving the capacity to revisit and render any past state of the dynamic scene. It is not an outlandish assumption that in application like sports and visual performance, we have multiview synchronized static camera setup. Therefore, any dynamic subject observed by multiple synchronized cameras only varies by a rigid transformation. Consequently, the function $F_t$ is approximated by a compact neural network:
\begin{equation}
F_t(\mathbf{x}, \mathbf{d}; \Theta_t) = \text{MLP}(\gamma(\mathbf{x}), \gamma(\mathbf{d}); \Theta_t),
\end{equation}
where $\gamma(\cdot)$ is the spatial hashing-based positional encoding and $\Theta_t$ are the learnable network parameters at time $t$.  By distributing the overall scene representation burden across multiple MLP over time, we achieve efficient dynamic scene modeling enabling time-archival capabilities. This design highlights a new direction for modular implicit neural representation, opening opportunities for scalability, compositionality, and retrospective neural rendering. Mathematically, 
\begin{equation}
    \mathcal{F}(\mathbf{x}, \mathbf{c}, t) = F_t(\gamma(\mathbf{x}), \gamma(\mathbf{d})); ~\textrm{where,} \gamma(.) ~\textrm{denotes positional encoding.}
\end{equation}
So, overall dynamic event can be written as a collection of $F_t$ or $\{F_1,  F_2,..., F_T \}$

\formattedparagraph{\textit{(iii)} Volume Rendering for View Synthesis.}
To synthesize novel views, we perform volumetric rendering along camera rays. Given a $n^\textrm{th}$ novel viewpoint parameterized by camera intrinsics $\mathbf{K}_n$ and extrinsics $(\mathbf{R}_n, \mathbf{t}_n)$, each pixel color is computed via numerical integration along rays cast into the volume:
\begin{equation}
\revisedone{
\hat{\mathbf{C}}(\mathbf{r}, t) = \int_{s_n}^{s_f} T^t(s) \sigma^t(\mathbf{r}(s)) \mathbf{c}^t(\mathbf{r}(s), \mathbf{d}) ds, \text{with} ~T^t(s) = \exp\left(-\int_{s_n}^{s} \sigma^t(\mathbf{r}(u)) du\right),}
\end{equation}
where $T^t(s)$ is the transmittance at time $t$, $\mathbf{r}(s) = \mathbf{o} + s\mathbf{d}$ defines the ray originating from camera center $\mathbf{o}$ in direction $\mathbf{d}$, parameterized by the distance $s$, with $s_n$ and $s_f$ indicating near and far clipping planes, respectively. Similar to \cite{muller2022instant}, we discretize this integral into $M$ samples for computational efficiency:
\begin{equation}
\revisedone{
\hat{\mathbf{C}}(\mathbf{r}, t) \approx \sum_{j=1}^{M} T_j^t(1 - e^{-\sigma_j \delta_j}) \mathbf{c}j, \quad T_j^t = e^{-\sum_{k=1}^{j-1} \sigma_k \delta_k},
}
\end{equation}
where $\delta_j$ is the distance between consecutive sampled points along the ray, and $(\mathbf{c}_j, \sigma_j)$ represent the color and density predictions at sample point $j$.

\formattedparagraph{\textit{(iv)} Training and Optimization.}
At each time instance $t$, the neural implicit function $F_t$ is trained using the captured multi-view images by minimizing the following photometric reconstruction loss:
\begin{equation}
\mathcal{L}(\Theta_t) = \sum_{i=1}^{N} \sum_{\mathbf{r} \in R_i} \left| \hat{\mathbf{C}}(\mathbf{r}; \Theta_t) - \mathbf{C}(\mathbf{r}) \right|_2^2 + \kappa \| \Theta_{t+1} - \Theta_t\|_2^2,
\end{equation}
where $\hat{\mathbf{C}}(\mathbf{r}; \Theta_t)$ is the rendered pixel color from viewpoint $i$ along ray $\mathbf{r}$, $\mathbf{C}(\mathbf{r})$ is the observed pixel color, and $R_i$ denotes the set of sampled rays from camera $i$, and $\kappa$ is a constant scalar. To facilitate temporal consistency across time instances, we optionally include temporal regularization terms or impose consistency constraints between consecutive neural implicit functions $(F_{t-1}, F_t)$. This encourages consecutive time MLPs to have similar weights. Yet, for maintaining the time efficiency while keeping the memory footprint as minimum as possible, in our experiment, we treats the dynamic event at each time step independently. Our take on this stems from the application this paper targets, i.e., sports and visual entertainment, where several dynamic subjects are present and their 3D positions are observed to change drastically between frames. Even empirically, independent time modeling has been observed to work well, and therefore, in this work we adhere to it. 

\formattedparagraph{\textit{(v)} Inference and Novel View Generation.}
At inference time, given a desired viewpoint not included in the original capture setup, we render novel views in real-time using the trained neural implicit functions. This design supports not only rendered image visualization but also retrospective rendering by querying archived scene representations $F_t$. As such, our approach enables time-aware visualizations that are particularly well-suited for sports replay, performance reenactment, and others.

\section{Experiment and Results}

\formattedparagraph{Implementation Details.} We implemented our approach using the PyTorch {2.5.1} and tested on NVIDIA GPUs with CUDA version 11.8.  All comparative experiments were performed on the NVIDIA A40 GPU (50 GB RAM), and additionally tested on the NVIDIA H100 GPU  (251 GB RAM). To enable efficient training and inference across temporally indexed neural scene representations, we adopted a modular design that supports time-wise synchronized optimization while preserving model parameters. We used synchronized calibrated camera intrinsic and extrinsic to evaluate our approach result. For benchmarking, we compared our approach against different state-of-the-art dynamic scene rendering methods, including D-NeRF \cite{pumarola2021d}, D-3DGS \cite{luiten2024dynamic}, 4DGS \cite{wu20244d}, and ST-GS \cite{li2024spacetime}. Quantitative and qualitative evaluations were performed on the publicly available CMU Panoptic Studio dataset \cite{joo2015panoptic}, which offers richly annotated multi-view video sequences suitable for dynamic scene analysis for complex human motion analysis. In addition, we evaluated performance on a newly introduced synthetic multiview dynamic scene dataset curated by us to reflect complex motion characteristics of visual performances and sports scenarios. This dual evaluation protocol ensures the robustness and generalizability of our method across diverse dynamic environments. We used the popular PSNR as well as the LPIPS metric to compare the image rendering quality with other methods, while the rendering speed is quantified using the FPS metric. More details on the model train time settings for each datasets---detailed next, are provided in \ref{Asubsec:model_training}. Code and dataset will be available at 
\href{https://drive.google.com/drive/folders/13n8dcC3czco81el9sSp8ZaDHLI0m9HLw}{link}.

\subsection{Dataset for Evaluation}
In this section, we first introduce our newly proposed dataset and experiment setup simulating the visual performance and sports events, which is the main focus of this paper. Next, we detail on our method's performance on CMU Panoptic dataset \cite{joo2015panoptic} and its experimental setup.

\formattedparagraph{Synthetic Multiview Dynamic Scene Dataset.}
We acquired synthetic datasets under three settings using Blender 4.0 for experimental evaluation: \textbf{\textit{(i)}} Dancing-Walking-Standing \textbf{\textit{(ii)}} Soccer Penalty Kick, and \textbf{\textit{(iii)}} Soccer Multiplayer. The 3D models used in creation of these datasets were taken from publicly available Internet assets.  

In our acquisition setup, we sample $N$ camera poses on a hemisphere of radius $R$ using Fibonacci sphere sampling. Given $N$ viewpoints ($i=0,1,\dots,N-1$), the $i$-th camera position \(\mathbf{p}_i\in\mathbb{R}^3\) is computed using the following relations between $\phi, \theta_i ~\text{and}~z_i$ (refer to \ref{Asubsec:fibo} for visual illustration):
\[
\phi     = \pi\,(3 - \sqrt{5}), \text{(the golden angle)};
\]
\[
 z_i = R \left(1 - \frac{i}{N}\right); ~\theta_i = (\phi \times i) \bmod 2\pi; ~r_i = \sqrt{R^2 - z_i^2};
~\mathbf{p}_i = \bigl(r_i\cos\theta_i,\;r_i\sin\theta_i,\;z_i\bigr).
\]


Here, \(R\) is the radius of the sphere. After obtaining the camera positions, we oriented each camera to face the scene center at \((0,0,0)\) and configured the intrinsics of the camera as follows:
\[
f_x = f_y = 2666.67,\quad
(c_x, c_y) = (960.0, 540.0),\quad
\text{resolution} = 1920 \times 1080.
\]
All radial distortion coefficients \(k_1 = k_2 = 0\) and tangential distortion coefficients \(p_1 = p_2 = 0\), implying an undistorted image. The resulting horizontal and vertical field of view angles are $\textrm{camera\_angle}_\textrm{x} = 0.6911\ \mathrm{rad},~\textrm{camera\_angle}_\textrm{y}  = 0.4711\ \mathrm{rad}$. Adopting synchronized multiview camera setup, at each discrete time instance, we acquire $\mathcal{I}_t$ (refer to Eq.\eqref{eq:1})

\noindent
\textit{\textbf{(i)} Dancing-Walking-Standing dataset.}  This dataset features three distinct dynamic subjects exhibiting complex motion patterns such as dancing, walking, and standing. A total of 65 time instances were captured, each rendered from 100 calibrated camera viewpoints.  We indexed $0^\textrm{th}, 30^\textrm{th}, 60^\textrm{th} $ and $90^\textrm{th}$ camera as our test cameras, while $1^\textrm{st}$ camera is used for validation. This leads to 95, 1, and 4 camera viewpoints for training, validation, and testing, respectively. This partitioning results in 6,175 training images ($65 \times 95$), 65 validation images ($65 \times 1$), and 260 test images ($65 \times 4$). All images are temporally synchronized, and the camera split is consistent across all time steps to ensure uniform view coverage and benchmarking reproducibility.

\noindent
\textit{\textbf{(ii)} Soccer Penalty Kick dataset.} This dataset consists of two dynamic subjects simulating a penalty kick scenario. We rendered 109 distinct time instances, each from 60 camera viewpoints. We indexed $21^\textrm{st}$, $37^\textrm{th}$, $40^\textrm{th}$, and $56^\textrm{th}$ camera as our test camera, while the $0^\textrm{th}$ camera is used for validation. This leads to 55 camera viewpoints for training, 1 for validation, and 4 for testing, providing us with 5,995 training images ($109 \times 55$), 109 validation images ($109 \times 1$), and 436 test images ($109 \times 4$). The cameras are aptly synchronized in time to capture diverse motion with consistent spatial sampling.

\noindent
\textit{\textbf{(iii)} Soccer Multiplayer dataset.} It captures three dynamic subjects in coordinated soccer actions. A total of 83 time instances were rendered using 60 camera viewpoints. Of these, 55 viewpoints were used for generating training images, 1 for validation, and 4 for testing. We indexed $21^\textrm{st}, 37^\textrm{th}, 40^\textrm{th}$ and $56^\textrm{th}$ camera number as our test camera, while $0^\textrm{th}$ camera is for validation. This results in 4,565 training images ($83 \times 55$), 83 validation images ($83 \times 1$), and 332 test images ($83 \times 4$). Similar to the previous dataset, the cameras are synchronized for proper evaluation. 

\begin{figure}[t]
\centering
\includegraphics[width=\textwidth]{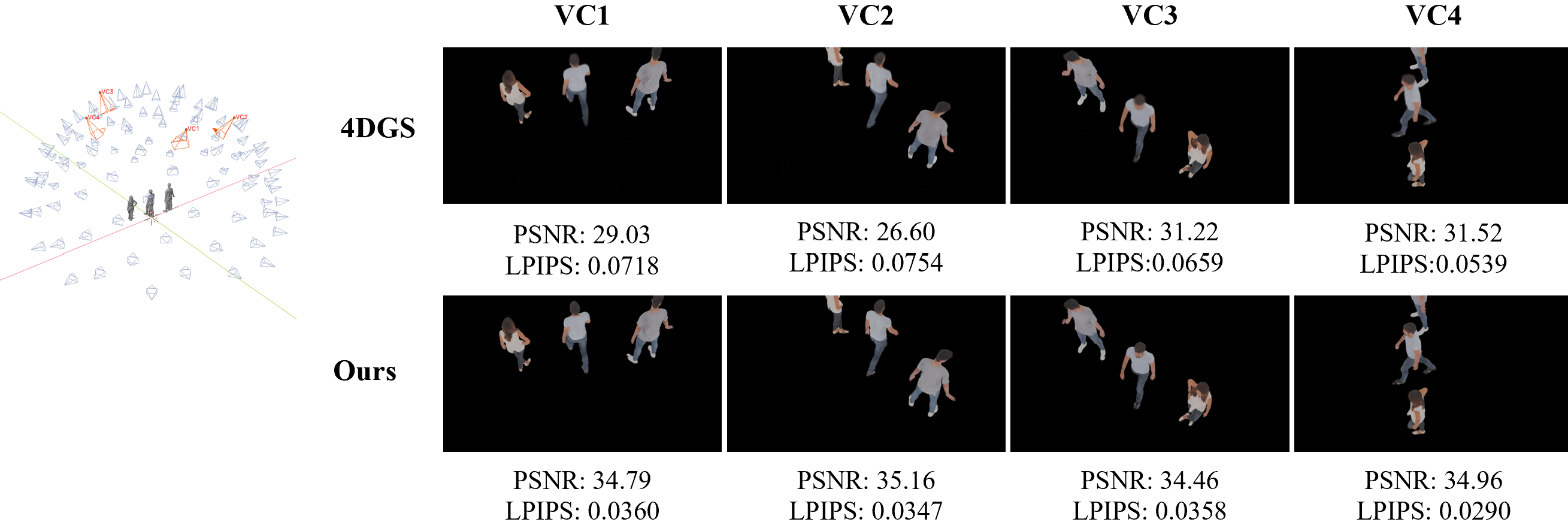}
\caption{\small Visual Performance Qualitative Comparison Results with 4DGS approach \cite{wu20244d} on our synthetic multiview dataset. \textbf{Left:} The four camera frustum highlighted in \textcolor{red}{red} shows the virtual cameras that will be used for dynamic scene broadcasting. \textbf{Right:} Our rendered image results from those virtual cameras at a given time as compared to 4D-GS \cite{wu20244d} approach. We also provide the PSNR and LPIPS values for quantitative comparison. Here, \textbf{VC} denotes corresponding virtual camera.}\label{fig:visual_comparison_synthetic}
\end{figure}

\begin{table}
\centering
\resizebox{0.9\textwidth}{!}
{
\begin{tabular}{lccccccc}
\toprule
{Method$\rightarrow$} & {{D-NeRF \cite{pumarola2021d}}} & {{D-3DGS \cite{luiten2024dynamic}}} & {{4DGS \cite{wu20244d}}} & {{ST-GS \cite{li2024spacetime}}} & \textbf{Ours} \\
\midrule
\colorbox{amber}{\textbf{Dancing-Walking-Standing}} & & & & & & \\
PSNR$\uparrow$ & 6.44 & 18.45 & 28.17 & 20.03  & \textbf{\colorbox{pastelgreen}{34.28}} \\
LPIPS$\downarrow$ & 0.572 & 0.139 & 0.08 & 0.112 & \textbf{\colorbox{pastelblue}{0.027}} \\
\midrule
\colorbox{amber}{\textbf{Soccer Penalty Kick}} & & & & & & \\
PSNR$\uparrow$ & 10.64 & 26.45 & 26.25 & 25.99  &  \textbf{\colorbox{pastelgreen}{33.81}} \\
LPIPS$\downarrow$ & 0.407 & 0.071 & 0.045 & 0.077 &  \textbf{\colorbox{pastelblue}{0.028}} \\
\midrule
\colorbox{amber}{\textbf{Soccer Multiplayer}} & & & & & & \\
PSNR$\uparrow$ & 6.15 & 26.43 & 26.2 & 25.92  &  \textbf{\colorbox{pastelgreen}{31.85}} \\
LPIPS$\downarrow$ & 0.533 & 0.087 & 0.061 & 0.104 &  \textbf{\colorbox{pastelblue}{0.039}} \\
\bottomrule 
\\
\end{tabular}
}
\caption{\small Quantitative comparison of our approach with state-of-the-art approaches on synthetic dataset. The best results are shaded with green and blue for PSNR and LPIPS metric, respectively}\label{tab:results_on_synthetic}
\end{table}

\formattedparagraph{Real World Multiview Dynamic Scene Dataset.} 
To evaluate our method on real-world dynamic scenes, we utilize the CMU Panoptic Studio dataset \cite{joo2015panoptic}. It offers densely captured multi-view human activity sequences. For controlled testing of dynamic subject rendering, we isolate the primary foreground subjects using high-precision semantic segmentation. Specifically, we apply the Segment Anything Model (SAM) \cite{kirillov2023segment} and its high-quality variant SAM-HQ \cite{ke2023segment} to extract clean subject masks for each frame. Refer to \ref{Asubsec:coordinate_conv} for details on coordinate conversion used for model training.

The preprocessing pipeline begins by detecting human subjects and associated props (e.g., baseball bats) in each frame using a YOLOv8 object detector. Detected bounding boxes are then passed to SAM-HQ to generate high-fidelity binary segmentation masks. These masks are aggregated into a single foreground mask per frame and used to convert the RGB images into RGBA format by embedding the merged mask into the alpha channel. This results in background-removed imagery that isolates dynamic subjects for controlled evaluation of dynamic subject rendering and evaluation.

We experiment on two challenging categories within the CMU Panoptic dataset. For the Baseball Bat sequence (1 subject), we extract 100 consecutive frames from the Sports1 subset, using 29 of the 31 high-definition cameras for training and the remaining 2 for testing. We used $10^\textrm{th}$ and $15^\textrm{th}$ camera as our test cameras. This configuration yields 2,900 training images and 200 test images. For the Hand Gesture sequence (1 subject), we extracted 201 consecutive frames from the Hands2 subset, following the same camera split, resulting in 5,829 training images and 402 test images. Such a controlled and high-quality pre-processing enables robust evaluation for real-world dynamic novel view synthesis methods.

\begin{figure}[t]
\centering
\includegraphics[width=0.86\textwidth]{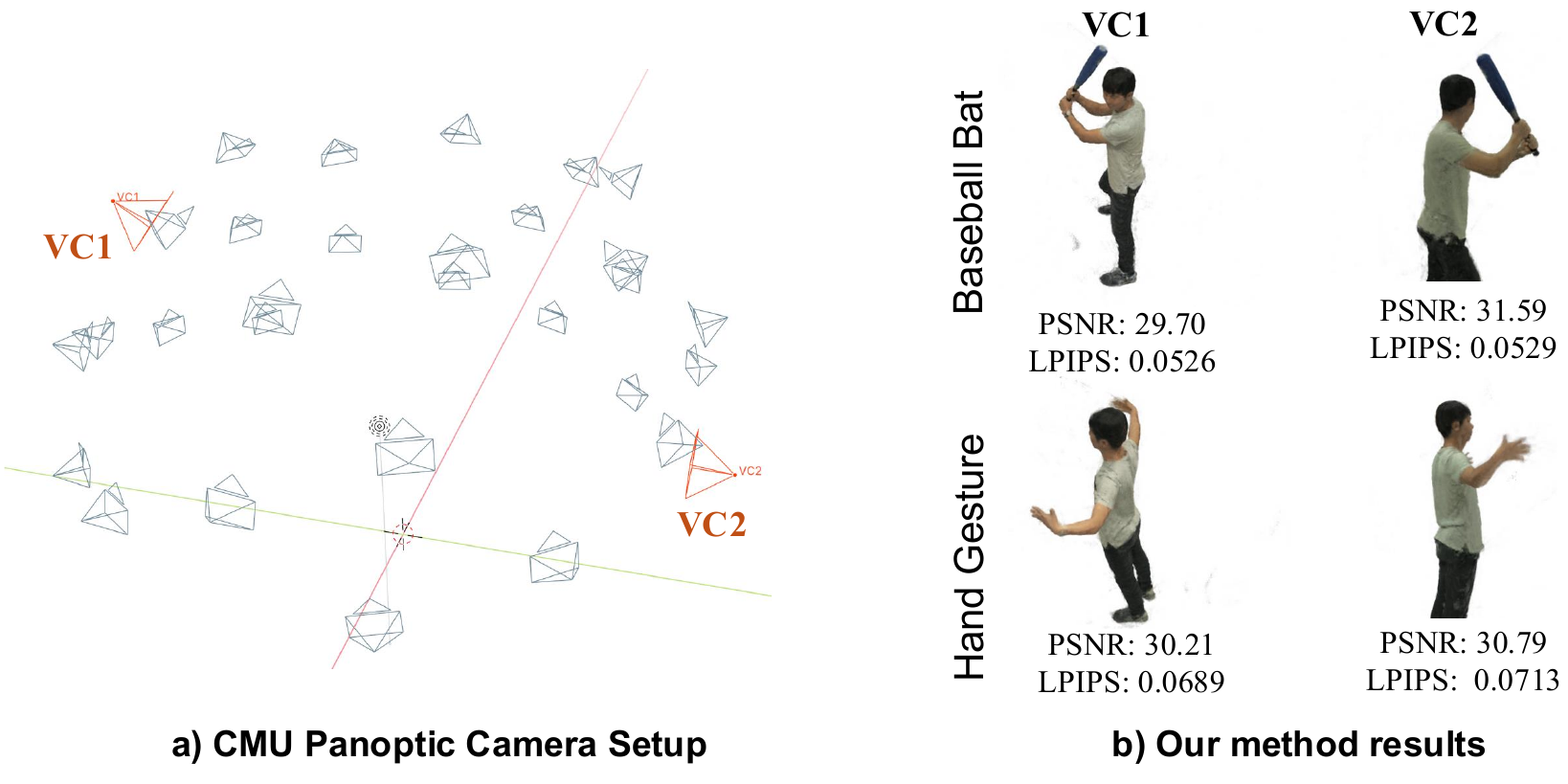}
\caption{\small Results on  CMU Panoptic dataset \cite{joo2015panoptic}. \textbf{Left:} Multiview Camera setup. Actual cameras are shown in black, where as virtual cameras are highlighted with \textcolor{red}{red}. \textbf{Right:} Results using our approach on a couple of challenging sports sequence. Here, \textbf{VC} denotes virtual camera.}\label{fig:visual_comparison_real}
\end{figure}

\begin{table}
\centering
\resizebox{0.9\textwidth}{!}
{
\begin{tabular}{lccccccc}
\toprule
{Method$\rightarrow$} & {{D-NeRF \cite{pumarola2021d}}} & {{D-3DGS \cite{luiten2024dynamic}}} & {{4DGS \cite{wu20244d}}} & {{ST-GS \cite{li2024spacetime}}} & \textbf{Ours} \\
\midrule
\colorbox{lavenderblue}{\textbf{Baseball Bat (CMU Panoptic \cite{joo2015panoptic, luiten2024dynamic} + SAM \cite{kirillov2023segment}) }} & & & & & & \\
PSNR$\uparrow$ & 6.35 & $\clubsuit$ & $\clubsuit$  & $\clubsuit$  & \textbf{\colorbox{pastelgreen}{29.43}} \\
LPIPS$\downarrow$ & 0.605 & $\clubsuit$ & $\clubsuit$ & $\clubsuit$ & \textbf{\colorbox{pastelblue}{0.066}} \\
\midrule
\colorbox{lavenderblue}{\textbf{HandGesture (CMU Panoptic \cite{joo2015panoptic, luiten2024dynamic} + SAM \cite{kirillov2023segment})}} & & & & & & \\
PSNR$\uparrow$ & 12.99 & $\clubsuit$ & $\clubsuit$ & $\clubsuit$  &  \textbf{\colorbox{pastelgreen}{29.19}} \\
LPIPS$\downarrow$ &0.135 & $\clubsuit$ & $\clubsuit$ & $\clubsuit$ &  \textbf{\colorbox{pastelblue}{0.050}} \\
\bottomrule 
\\
\end{tabular}
}
\caption{\small Quantitative comparison of our approach with state-of-the-art approaches on CMU Panoptic dataset. $\clubsuit$ symbolizes that the corresponding method fail to give results. Note that D-3DGS \cite{luiten2024dynamic} uses active sensor depth data, which is not present in our experimental setting. }\label{tab:results_on_real}
\end{table}

\subsection{Evaluation and Result}
We primarily assess our experimental results using various metrics. These include the peak-signal-to-noise ratio (PSNR) and the perceptual quality measure LPIPS \cite{zhang2018unreasonable}, which measures the quality of rendered images. We further analyze the rendering speed using the popular FPS metric.

\formattedparagraph{\textit{(i)} Results on Synthetic Multiview Dynamic Scene Dataset.} We compared our approach results with several state-of-the-art methods, such as D-NeRF \cite{pumarola2021d}, D-3DGS \cite{luiten2024dynamic}, 4DGS \cite{wu20244d}, and SpaceTime GS \cite{li2024spacetime}, to examine our performance on dynamic scene novel view synthesis.  Table \ref{tab:results_on_synthetic} provides the quantitative results obtained on the proposed synthetic dynamic scene dataset. While the state-of-the-art method shows some promise, it generally fails to render convincing results on the proposed dynamic scene. Moreover, the notion of time archival is absent in all the previous methods, hence unsuitable for sports and visual performance applications. Figure \ref{fig:visual_comparison_synthetic} shows a qualitative comparison of our approach with 4DGS \cite{wu20244d}, clearly demonstrating the benefit of our approach.

\formattedparagraph{\textit{(ii)} Results on CMU Panoptic Dataset.} Table \ref{tab:results_on_real} presents the results obtained from real-world datasets. D-3DGS need depth data, which is not an input for the experimental setup, thus shown with $\clubsuit$ symbol. Furthermore, it is apparent that 4DGS \cite{wu20244d} and  ST-GS \cite{li2024spacetime} will fail due to over-reliance on 3D points priors from the structure-from-motion framework. On the contrary, our method successfully provides dynamic scene rendering results at the current time. It also provides the flexibility to analyze the same scene in the past and allows the user to place the new virtual camera and examine the scene retrospectively. Surprisingly D-NeRF \cite{pumarola2021d}  provides unsatisfactory results and this maybe due to the linear deformation field approximation in their approach. Figure \ref{fig:visual_comparison_real} provides qualitative results obtained on this dataset using our method.

{\centering
\begin{minipage}{0.5\textwidth}
  \small
  \begin{tabular}{lcccccc}
    \toprule
    {Method$\rightarrow$} &  {{4DGS \cite{wu20244d}}} & {{ST-GS \cite{li2024spacetime}}} & \textbf{Ours}\\
    \midrule
    PSNR$\uparrow$ &  17.08 & 20.03 &  \textbf{34.28} \\
    LPIPS$\downarrow$ &  0.123 & 0.112 & \textbf{0.027} \\
    \bottomrule
    \end{tabular}
  \captionof{table}{\small The proposed approach faithfully represent a dynamic scene compared to 3DGS based extensions. The above results for other methods are obtained on our synthetic Dancing-Walking-Standing sequence with random 3d point initialization.}\label{tab:ablation_results_random_3d}
\end{minipage}
\quad
\begin{minipage}{0.45\textwidth}
  \centering
  \includegraphics[width=0.85\textwidth]{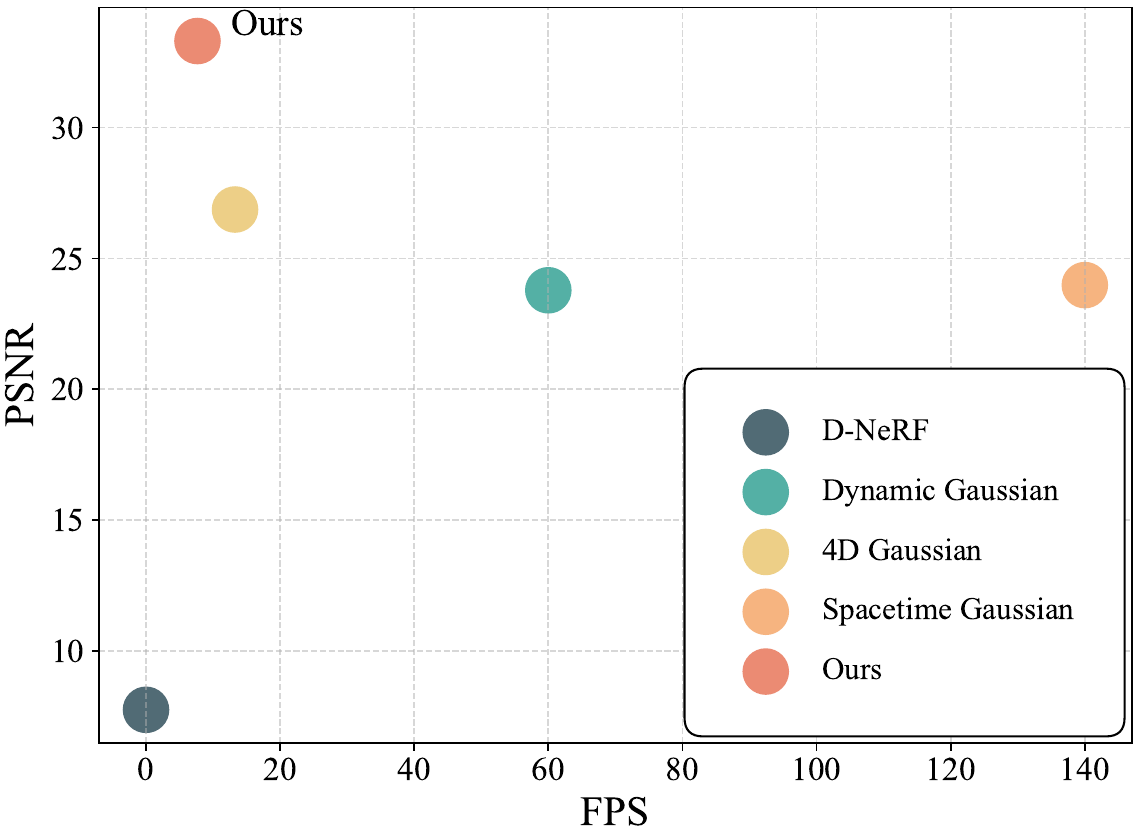}
  \captionof{figure}{\small Quantitative comparison of the FPS and PSNR on the proposed dynamic scene dataset.}\label{fig:psnr_fps_comparison}
\end{minipage}
}

\subsection{Ablations}
\formattedparagraph{\textit{(i)} Random 3D points initialization for 3DGS based approaches.} As shown in Table \ref{tab:results_on_real} that 3DGS \cite{kerbl20233d} based methods for dynamic scene novel view synthesis, such as 4DGS \cite{wu20244d} and ST-GS \cite{li2024spacetime}, fails on real world scene. This is due to over-reliance on accurate 3D points from SfM to anchor the spatial distribution of Gaussian primitives. By allowing flexibility to these methods assuming a less ideal condition, we initialize Gaussian centers by randomly sampling 3D positions within the bounding volume of the scene. The hope is that such random initialization could serve as a warm start for 3DGS based approaches to initiate the optimization process for favorable image rasterization.

The objective of this ablation was to assess whether current 3DGS frameworks possess sufficient representational and optimization flexibility for a complex dynamic scene. Empirical results when tested on synthetic Dancing-Walking-Standing dataset (see Table \ref{tab:ablation_results_random_3d}) clearly show their limitation for dynamic scene applications compared to ours that leverages rigidity across synchronous camera views for neural scene representation learning without any requirement of 3d points from SfM pipelines.

\formattedparagraph{\textit{(ii)} Trade-off Between Rendering Speed and Quality in Dynamic Novel View Synthesis.} The objective is to analyze the metric between frames-per-second (FPS) and image rendering quality (measured in PSNR). Figure \ref{fig:psnr_fps_comparison} provides the quantitative results for the same. While 3DGS-based approaches, such as 4D-GS \cite{wu20244d} and ST-GS \cite{li2024spacetime}, deliver high-speed novel view synthesis results via forward rasterization of explicit primitives, our work observed a notable limitation in their applicability to dynamic scenes. Specifically, despite 3DGS methods achieve high FPS, their rendering quality degrades in the presence of complex dynamic motion, often yielding unsatisfactory PSNR scores. 

In contrast, our proposed implicit neural scene representation framework provides a more favorable balance between quality and efficiency. By learning temporally indexed radiance fields through a compact yet expressive neural architecture, our method achieves consistently high PSNR across challenging dynamic scenes while maintaining real-time or near-real-time FPS on modern GPU hardware. These results highlight the limitations of relying solely on speed-centric splatting methods for dynamic environments. We position our approach as a more acceptable solution for the targeted applications requiring temporal consistency and photorealistic rendering quality. Yet, our current implementation provides 4-5 FPS, i.e., near real-time performance, a \textbf{limitation} nonetheless\footnote{refer to Appendix for more experimental analysis and video}.

\begin{table}[t]
\resizebox{0.96\textwidth}{!}
{
\begin{tabular}{|c|c|c|c|c|}
\hline
\textbf{Approach/Evaluation Metric} & \textbf{PSNR} & \textbf{LPIPS} & \textbf{Model} (per time step) & \textbf{Input PointCloud} (per time step) \\ \hline
3DGS (GT point cloud) &  36.47    &   0.0916  &  77 MB & $\sim$6.2 GB \\ \hline
3DGS (random point cloud) &  16.33   & 0.3761 & 91 MB & $\sim$6.2 GB \\ \hline
Ours (no point cloud) &  \textcolor{blue}{\textbf{34.28}} &   \textbf{0.0255} & \textbf{48.8} MB & \textbf{0.0} 
\\ \hline
\end{tabular}
}
\caption{\small  \revisedone{Quantitative comparison of our approach with 3DGS  for single time instance. Without any ground truth 3d point cloud priors, our approach provides excellent results with significantly low-memory requirement. On the contrary, if we use 3DGS approach here instead of the proposed implicit approach for each time step, we need to maintain a persistent dense folder with 3d data by possibly running COLMAP for each time step (due to dynamic scene setup). For the Dance-Walking-Standing dataset, this folder occupies nearly 6.2 GB to achieve the results mentioned above.} }\label{tab:each_instance_3dgsvsours}
\end{table}

\formattedparagraph{\revisedone{\textit{(iii)} 3DGS vs Our MLP based representation on Single Time Instance.}} \revisedone{To further understand the suitability of implicit versus explicit scene representations for time-archival in dynamic sports and visual-performance settings, we conducted a controlled single-time-instance ablation comparing our per-time-step MLP-based radiance field with 3D Gaussian Splatting under two initialization regimes on Dance-Walking-Standing dataset: (i) using an accurate ground-truth (GT) point cloud, and (ii) using a randomly initialized point cloud. The results in Table~\ref{tab:each_instance_3dgsvsours} reveal several important insights. First, 3DGS achieves strong performance only when provided with a high-quality GT point cloud (PSNR 36.47, LPIPS 0.0916), confirming its reliance on precise geometric priors for stable reconstruction. However, in the absence of such priors---which is the realistic setting for fast-paced sports scenes with severe articulation, occlusion, or motion blur—the performance of 3DGS degrades drastically (PSNR 16.33, LPIPS 0.3761). In contrast, our implicit representation, which requires \textbf{no} point-cloud initialization, produces consistently high-quality renderings (PSNR 34.28, LPIPS 0.0255) comparable to 3DGS with GT points but without any geometric supervision. Moreover, the memory footprint per time step for our compact radiance field (48.8 MB) is \textbf{significantly smaller} than that of 3DGS (77–91 MB), highlighting a critical advantage for \textbf{long-horizon time-archival}. Furthermore, our approach overcomes the limitation of maintaining the persistent 3d data for 3DGS to work, saving nearly 6.2 GB of memory on the test case presented in Table~\ref{tab:each_instance_3dgsvsours}. Sports broadcasts and performance recordings often span hundreds or thousands of frames, and storing explicit Gaussian fields per time step becomes prohibitively expensive and brittle due to their dependence on accurate per-frame geometry. This ablation demonstrates that our implicit formulation is not only more robust in the absence of high-quality 3D geometry but also far more storage-efficient, making it particularly well suited for scalable time-archival of dynamic events where extreme motion, rapid pose changes, and multi-actor interactions undermine the assumptions required by Gaussian-based representations.}

\begin{table}[t]
\resizebox{0.96\textwidth}{!}
{
\begin{tabular}{|c|c|c|c|c|}
\hline
Method & Train Time & \begin{tabular}[c]{@{}c@{}}Parallelizable \\ (in Time)\end{tabular} & PSNR & LPIPS \\ \hline
4DGS \cite{wu20244d} & $\sim 0.30 - 3.0$ hours & No (sequential) & $\sim$ 26.2 - 28.1 & $\sim$ 0.045 - 0.061 \\ \hline
ST-GS \cite{li2024spacetime} & $\sim 0.46 - 0.71$ hours & No (sequential) & $\sim$ 20.0 - 25.9 & $\sim$ 0.078 - 0.112 \\ \hline
D-3DGS \cite{luiten2024dynamic} & $\sim 1.40$ hours & No (Limited) & $\sim$ 18.5 - 26.5 & $\sim$ 0.071 - 0.139 \\ \hline
Ours   & $\sim$ 5.65 - 8.90  hours &  \textbf{Yes (Fully)} & $\sim$ \textbf{31.9 - 34.3} & $\sim$ \textbf{0.027 - 0.039} \\ \hline
\end{tabular}
}
\caption{\small \revisedone{Our per-time-step radiance fields achieve the highest reconstruction quality (PSNR, LPIPS) while offering fully parallelizable training across time, unlike sequential 4DGS, ST-GS, and D-3DGS based approaches. Despite a higher per-sequence training cost on a single GPU, full parallelization makes our method significantly more scalable for long-horizon time-archival in sports and visual-performance applications.} }\label{tab:train-time-comparison-4DGS}
\end{table}

\formattedparagraph{\revisedone{\textit{(iv)}  Full pipeline train-time and result comparison compared to 4DGS and other similar baselines.}}
\revisedone{To further assess the practical feasibility of our per-time-step neural representation for time-archival applications in sports and visual performance, we compare the end-to-end training time of our method against state-of-the-art dynamic Gaussian splatting pipelines and dynamic NeRF variants. As summarized in Table~\ref{tab:train-time-comparison-4DGS}, methods such as 4DGS~\cite{wu20244d}, ST-GS~\cite{li2024spacetime}, and D-3DGS~\cite{luiten2024dynamic} are inherently sequential, i.e., their optimization relies on propagating Gaussian primitives, motion fields, or deformation information across frames, making them unsuitable for fully parallel processing of long dynamic sequences. Their per-sequence train-time consequently ranges from approximately 0.3 to 3.0 hours for 4DGS and up to 1.4 hours for D-3DGS. In contrast, although our radiance-field models require 5.65–8.90 hours of total training time for an entire sequence on a single NVIDIA A 6000 GPU, each time step is \emph{independent}, making our approach trivially parallelizable across tens or hundreds of GPUs. This parallelism capability---absent in Gaussian-based dynamic methods---effectively reduces wall-clock training time by an order of magnitude in multi-GPU settings, which is critical for practical deployment in broadcast or replay pipelines where rapid turnaround is essential. Importantly, our fully parallelizable design does not sacrifice reconstruction quality and achieves the highest PSNR (31.9–34.3) and lowest LPIPS (0.027–0.039) across all baselines, substantially outperforming Gaussian-based methods for dynamic scene. These results demonstrate that independent, per-time-step implicit radiance fields provide a compelling trade-off between compute, accuracy, and scalability, particularly in the context of time-archival for dynamic sports scenes, where long sequences, rapid motion, and high-fidelity reconstruction requirements make sequential Gaussian-based pipelines prohibitive.}

\begin{table}
\centering
\resizebox{\textwidth}{!}
{
\begin{tabular}{l|>{\columncolor{blue!15}}c|>{\columncolor{blue!15}}c|>{\columncolor{blue!15}}c|>{\columncolor{red!15}}c|>{\columncolor{red!15}}c|>{\columncolor{red!15}}c|>{\columncolor{red!15}}c|>{\columncolor{red!15}}c|>{\columncolor{orange!40}}c}
\toprule
{Method$\rightarrow$} & 
{{D-3DGS \cite{luiten2024dynamic}}} & 
{{4DGS \cite{wu20244d}}} & 
{{ST-GS \cite{li2024spacetime}}} &
{{D-NeRF \cite{pumarola2021d}}} &
{{T-4D \cite{shao2023tensor4d}}} & 
{{HP \cite{cao2023hexplane}}} & 
{{KP \cite{fridovich2023k}}} & 
{{S-RF \cite{li2022streaming}}} &
\textbf{Ours} \\
\midrule

{\textbf{DWS}} & & & & & & & & & \\
\hline
PSNR    & 18.45 & 28.17 & 20.03 & 6.44  & 16.55 & 15.82 & 16.40 & 18.87 & \textbf{34.28} \\
LPIPS   & 0.1396 & 0.0800 & 0.1120 & 0.5726 & 0.2165 & 0.2470 & 0.2327 & 0.2054 & \textbf{0.0275} \\
Memory  & 2.0MB & 21.0MB & 3.2MB & 512k  & 280MB & 68MB & 419MB & 2.2GB & 3.1GB \\
Train Time & 1.41h & 0.83h & 0.46h & 4.25h & 10.02h & 0.96h & 1.02h & 0.65h & 5.25h \\
Iterations & 130K & 17K & 30K & 20K & 200K & 25K & 30K & 80K & 19K \\
\midrule

{\textbf{S-PK}} & & & & & & & & & \\
\hline
PSNR    & 26.45 & 26.25 & 25.99 & 10.64 & 25.86 & 22.45 & 21.30 & 22.28 & \textbf{33.81} \\
LPIPS   & 0.0719 & 0.0450 & 0.0778 & 0.4070 & 0.0588 & 0.1567 & 0.1866 & 0.1792 & \textbf{0.0282} \\
Memory  & 2.0MB & 13.9MB & 0.9MB & 197K & 280MB & 68MB & 419MB & 1.8 GB & 5.2GB \\
Train Time & 1.48h & 3.02h & 0.47h & 7.74h & 9.12h & 0.75h & 0.94h & 0.55h & 8.90h \\
Iterations & 130K & 17K & 30K & 20K & 200K & 25K & 30K & 80K & 16.5K \\
\midrule

{\textbf{S-MP}} & & & & & & & & & \\
\hline
PSNR    & 26.43 & 26.20 & 25.92 & 6.15 & 25.82 & 20.42 & 19.34 & 24.98 & \textbf{31.85} \\
LPIPS   & 0.0872 & 0.061 & 0.104 & 0.5330 & 0.0881 & 0.2104 & 0.2115 & 0.0906 & \textbf{0.0392} \\
Memory  & 2.0MB & 19MB & 0.9MB & 228K & 280MB & 68M & 419MB & 1.8GB & 4.0GB \\
Train Time & 1.46h & 2.94h & 0.71h & 4.21h & 10.1h & 0.83h & 1.03h & 0.58h & 6.28h \\
Iterations & 130K & 17K & 30K & 20K & 200K & 25K & 30K & 80K & 16.5K \\
\bottomrule
\end{tabular}
}
\caption{\revisedone{\small Comparison of implicit and explicit dynamic-scene representations on three synthetic datasets. Our per-time-step implicit radiance fields achieve the highest PSNR/LPIPS performance while maintaining a compact memory footprint, outperforming 3DGS- and NeRF-based methods that rely on explicit geometry or deformation tracking. This makes our approach substantially more scalable and robust for long-horizon time-archival in dynamic sports and performance scenes. \emph{Note: here we have not included the initial 3d point cloud size that is used by 3DGS based approaches as input}. \textbf{DWS}, \textbf{S-PK}, and \textbf{S-MP} stand for Dance-Walking-Standing, Soccer Penalty-Kick, and Soccer Multi-Player dataset.}}
\label{tab:comparison_result_with_both_3dgs_and_nerf_approaches}
\end{table}

\begin{figure*}[t]
\centering
\includegraphics[width=1.0\textwidth]{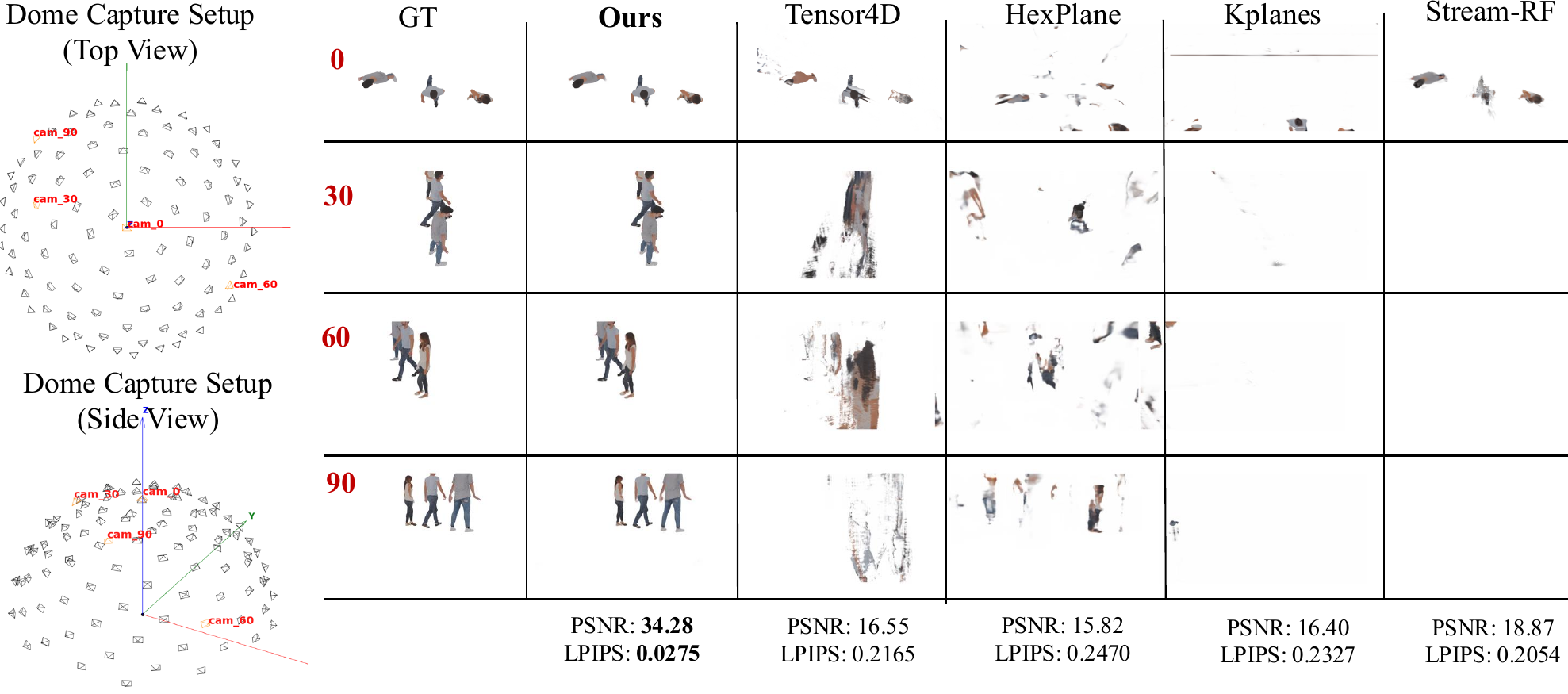}
\caption{\small \revisedone{Qualitative results demonstrating dynamic scene view synthesis from novel viewpoints on our proposed Dancing-Walking-Standing dataset. \textbf{Left:} shows the acquisition setup from top view and side view, while the novel views are shown in red. \textbf{From Top row to Bottom row}: Image-based rendering result with the state-of-the-art neural implicit methods for dynamic scenes. The red numbers in each row indicate the camera-id used by the methods to render the scene. Our approach generates images closer to the ground truth than others.}} \label{fig:visual_comparison_with_implicit_methods}
\end{figure*}

\begin{figure}[t]
\centering
\includegraphics[width=1.0\textwidth]{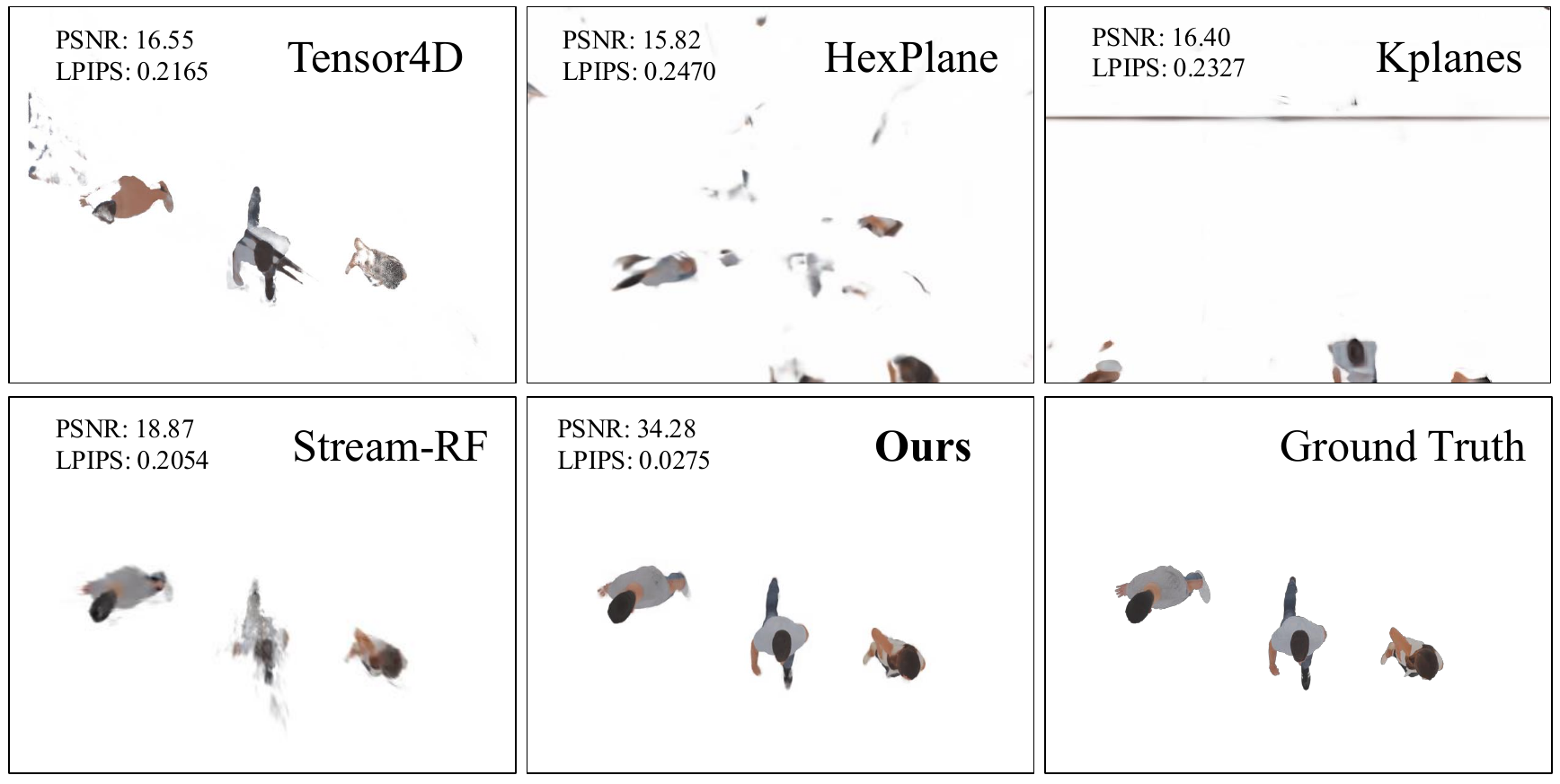}
\caption{\small \revisedone{Additional qualitative results showing results using different approaches on Dancing-Walking-Standing dataset.} }\label{fig:visual_comparison_with_implicit_methods-1}
\end{figure}

\formattedparagraph{\revisedone{\textit{(v)} Comparison with both implicit and explicit representation baselines for dynamic scene time-archival view synthesis.}}
\revisedone{To assess the suitability of implicit versus explicit scene representations for time-archival view synthesis in dynamic environments, we conducted a comprehensive comparison against both 3D Gaussian Splatting (3DGS)–based dynamic methods and recent implicit spatio-temporal radiance-field models (Table~\ref{tab:comparison_result_with_both_3dgs_and_nerf_approaches}). Across all three synthetic datasets, namely, Dance-Walking-Standing (DWS), Soccer Penalty-Kick (S-PK), and Soccer Multi-Player (S-MP) our approach consistently achieves the highest reconstruction quality, outperforming 3DGS variants and dynamic NeRF baselines by a substantial margin in both PSNR and LPIPS. Notably, dynamic splatting methods such as D-3DGS, 4DGS, and ST-GS exhibit a strong dependence on initialization quality and temporal regularization, which leads to performance degradation in scenes with rapid articulation, multi-subject interactions, and occlusion patterns common in sports. Implicit factorization approaches (Tensor4D, HexPlane, K-Planes, and StreamRF) provide smoother reconstructions but struggle to maintain fidelity under abrupt motion discontinuities---refer to Figure \ref{fig:visual_comparison_with_implicit_methods} and Figure \ref{fig:visual_comparison_with_implicit_methods-1} for visual comparison on Dancing-Walking-Standing dataset. In contrast, our per-time-step implicit radiance fields remain stable even under extreme motion, owing to their independence from deformation graphs and Gaussian tracking. Crucially, while 3DGS-based approaches often require large explicit point or Gaussian sets, our method stores only a compact MLP per time step, yielding a significantly more scalable representation for long-horizon temporal archival. This computational efficiency is especially important for sports broadcasting and performance capture, where hundreds or thousands of frames must be retained and queried retrospectively. Taken together, these results demonstrate that our implicit, temporally indexed formulation provides a more robust, memory-efficient, and high-fidelity solution for dynamic scene time-archival compared to both explicit splatting-based and implicit factorized alternatives.}

\begin{wrapfigure}{R}{0.40\textwidth}
\centering
\includegraphics[width=0.40\textwidth]{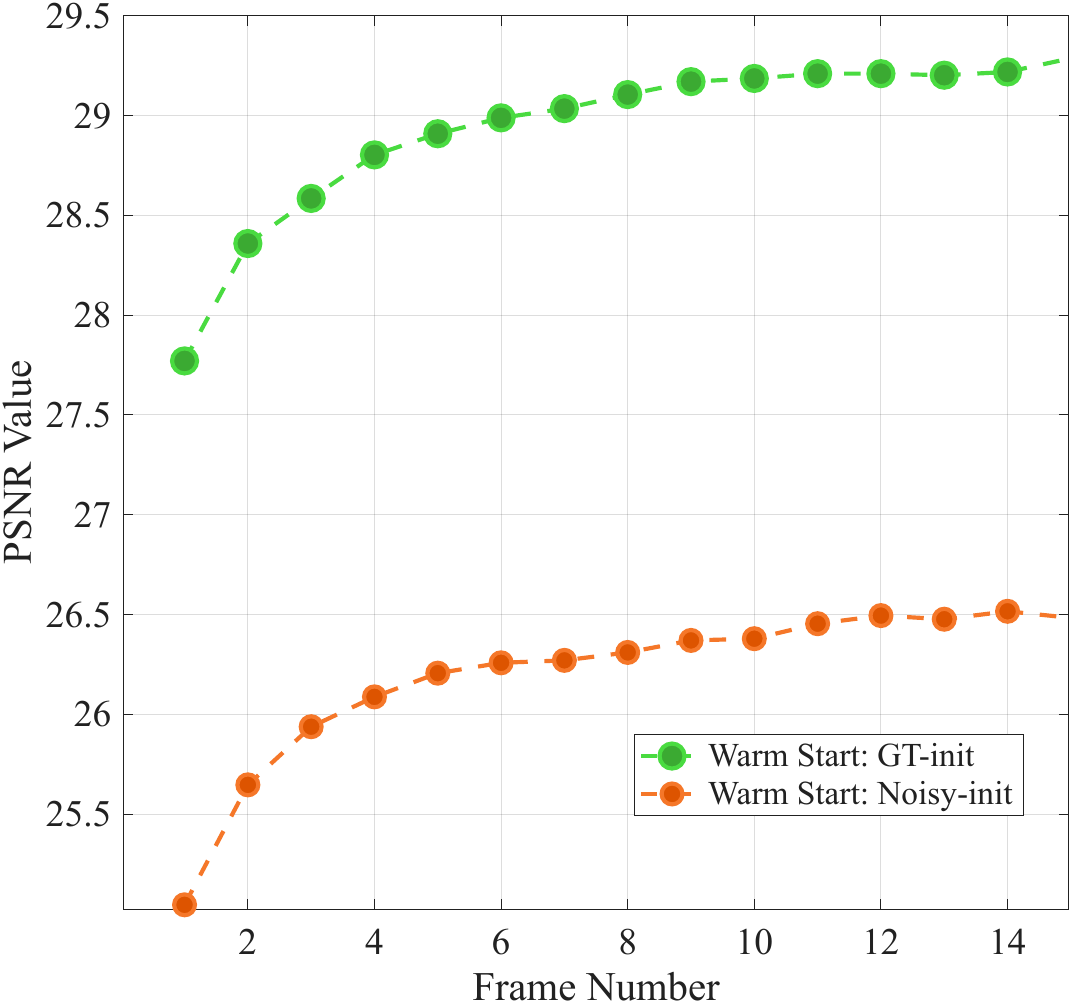}
\caption{\small \label{fig:warm-up-init-experiment} Per-frame PSNR variation, illustrating the sensitivity of 3DGS-based approaches to the quality of the reference 3D point cloud (warm start).}
\end{wrapfigure}

\formattedparagraph{\secondrevisedone{\textit{(vi)} Warm-Start Chaining in Time-Archival (Per-Timestep) 3D Gaussian Splatting.}}
\secondrevisedone{
By Warm-Start, we mean a good point cloud initialization to start dynamic Gaussian Splatting\footnote{Note however that the term dynamic Gaussian Splatting here implies the per-timestep (time-archival) 3DGS.} approach. To evaluate the robustness of warm-start chaining in dynamic Gaussian Splatting, where the optimized model at time step $t$ is used to initialize time step $t+1$, we conducted a controlled ablation on a 15-frame soccer sequence under two settings. All experimental settings were kept identical across runs, including warm-start chaining, a fixed Gaussian count (densification disabled), 8000 optimization iterations per frame, and identical rendering parameters $(r=1)$. The only difference between the two runs was the quality of the point-cloud initialization at the first frame.}

\secondrevisedone{In the GT-initialization (GT-init) setting, the first frame was initialized using a Blender-exported ground-truth point cloud containing approximately 100k points. In the Noisy-initialization (Noise-init) setting, we constructed a deliberately degraded initialization by retaining only $1\%$ of the original points and replacing the remaining $99\%$ with uniformly sampled points within an expanded bounding box (bbox\_scale = 4.0), with randomized RGB values, while keeping the total point count unchanged. Despite identical warm-start chaining thereafter, the GT-initializtion run achieved PSNR 28.94/SSIM 0.851/LPIPS 0.390, whereas the Noisy-initialization run dropped to PSNR 26.20/SSIM 0.812/LPIPS 0.430. Notably, the resulting $\sim 2.74$dB PSNR gap persisted consistently across frames 1–15, indicating that inaccuracies in the first-frame initialization propagate through the entire warm-start chain and are not substantially corrected by subsequent optimization.}

\secondrevisedone{These results demonstrate that, while warm-start chaining can improve efficiency when accurate geometry is available, its performance---contrary to ours---remains strongly dependent on the accuracy of the reference-frame point cloud. In dynamic sports scenarios, where occlusion, motion blur, and rapid articulation frequently degrade point-cloud quality thus posing limitation and motivates use of self-contained per-time-step implicit representations for reliable dynamic scene time-archival.
}

\section{\secondrevisedone{Discussion}}

\secondrevisedone{\textbf{Independence Vs Temporal Coupling}. An important design decision in this work is to model each time step with an independent implicit radiance field, rather than enforcing explicit information sharing across time steps. While many dynamic-scene methods, both implicit and Gaussian-based, successfully exploit temporal coupling to improve efficiency and coherence, such coupling implicitly assumes smooth motion, stable correspondences, or persistent primitives across frames. In the sports and visual-performance scenarios we target, these assumptions are frequently violated due to abrupt motion, strong articulation, multi-person interactions, and rapid occlusion changes. In such regimes, temporal coupling can become brittle, leading to drift or bias accumulation. By contrast, treating each time step as an independent, geometry-constrained reconstruction problem enables exact temporal indexing and drift-free archival, which is central to retrospective analysis and replay. Our decision to train independent models per time step is therefore not meant to suggest that Gaussian-based or temporally coupled methods are inherently inferior. Instead, it reflects a deliberate design choice aligned with the specific problem setting we target, i.e., time-archival camera virtualization under synchronized multi-view capture, \emph{where each time step is already strongly constrained by geometry} and where extreme, non-smooth motion and multi-actor interactions can make temporal coupling brittle. In these regimes, enforcing information exchange across time may introduce drift or bias, whereas independent optimization provides exact, temporally indexed reconstructions.}

\section{\secondrevisedone{Limitations}}
\secondrevisedone{
While the proposed time-archival camera virtualization framework demonstrates strong performance on dynamic sports and visual-performance  scenes captured with synchronized multi-view capture setups, the proposed framework has limitations. Firstly, our per-time-step formulation enables straightforward parallelization across time, yet \emph{full} parallelism of our approach could require additional GPU resources which may not always be available in practice. On the other hand, while some components of dynamic Gaussian splatting can also be parallelized, many dynamic Gaussian splatting pipelines remain partially sequential in practice c.f., Table \ref{tab:train-time-comparison-4DGS}. This is due to temporal dependencies such as Gaussian propagation, deformation tracking, or canonical-space optimization, which limit the extent of parallelization without introducing approximations. Secondly, small fine grained low-lying object(s) on ground or stages (such as grass, bottles) \cite{kumar2012bayes, mittal2014small} and challenging illumination conditions (such as mixed color temperatures, specular highlights, rapid exposure changes, or complex stage lighting) \cite{yaoreflective, jiang2024gaussianshader} remains boundary case. Under these conditions, we occasionally observe artifacts such as faint color leakage or reduced contrast in the rendered results. In practice, incorporating an explicit foreground–background separation stage (e.g., SAM-HQ \cite{kirillov2023segment, ke2023segment} or similar high-quality segmentation) could help mitigate these effects, while more explicit illumination modeling is an important direction for future work.}

\section{Conclusion}
In this paper, we introduced an approach for time-archival camera virtualization in dynamic scenes, unlike \cite{li2020crowdsampling, liu2020learning}, targeting applications in sports and visual performance. Our approach learns temporally indexed neural implicit representations from synchronized multi-view inputs, enabling photorealistic novel view synthesis across space and time. Unlike 3DGS-based extensions to dynamic scenes that rely on explicit geometry and fail under complex motion, our approach offers superior temporal consistency and fidelity. A key contribution is our ability to implicitly model the plenoptic function without requiring explicit 3D points as input as the subject(s) in the scene move through time. This enables re-rendering from arbitrary viewpoints at any past moment, supporting archival and interactive replay. Extensive evaluations demonstrate high-quality image rendering, outperforming current state-of-the-art baselines in targeted areas of application, such as sports and visual performance.

\bibliographystyle{elsarticle-num}
\bibliography{main}

\appendix
\section{Technical Appendices}

\subsection{Synthetic Multiview Dataset Camera Pose Acquisition via Fibonacci Sphere Sampling}\label{Asubsec:fibo}

For our synthetic multiview dataset, we assumed uniformly distributed camera setup. The camera positions are assigned uniformly over the surface of a hemisphere using Fibonacci sphere sampling. Our approach ensures comprehensive coverage and approximately equal angular spacing between camera viewpoints. For a given hemisphere of radius \( R \), the method calculates positions for \( N \) camera viewpoints indexed by \( i \),~$\forall$~\( i \in \{0, 1, \dots, N-1\} \).

\begin{figure}
    \centering
    \includegraphics[width=0.7\linewidth]{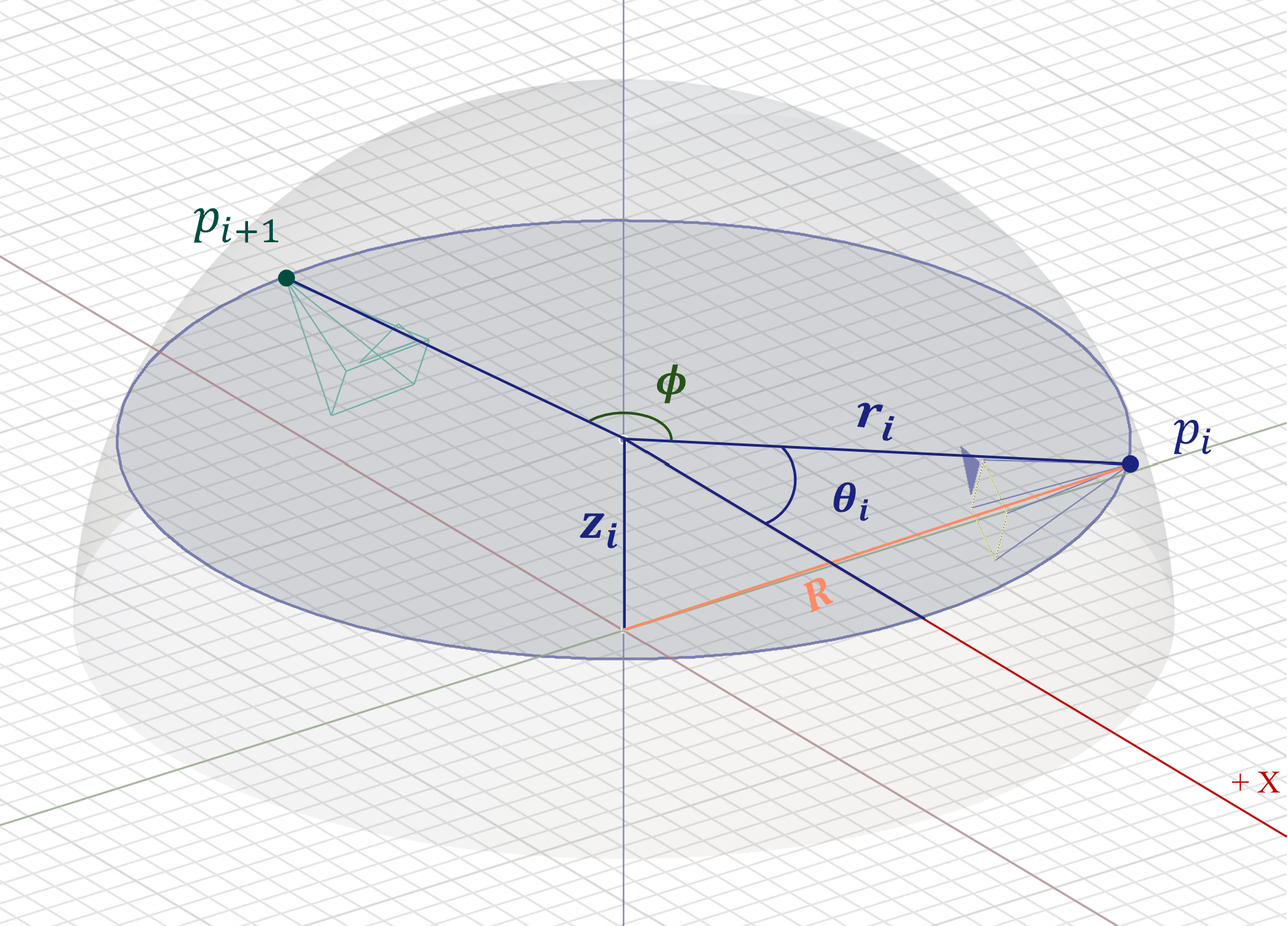}
    \caption[Hemisphere Fibonacci sampling]{
    Fibonacci sampling on a hemisphere of radius \(R\) for camera placement.
    The \(i\)-th camera center \(\mathbf p_i\) (dark-blue solid marker) has vertical coordinate \(z_i\),
    in-plane radius \(r_i=\sqrt{R^2-z_i^2}\), and azimuth \(\theta_i\).
    The next center \(\mathbf p_{i+1}\) (green solid marker) is rotated by the golden
    angle \(\phi\) about the vertical axis and lies at height \(z_{i+1}\!\approx\!z_i\).}
    \label{fig:fibonacci_sphere_sampling}
\end{figure}

The camera position \(\mathbf{p}_i\) for the \( i \)-th viewpoint is computed via following steps---refer to Figure \ref{fig:fibonacci_sphere_sampling} for visual  understanding of the camera position computation in our experimental setup.

\begin{enumerate}
    \item \textbf{Golden Angle}: We utilize the golden angle \(\phi \approx 2.39996\) radians, defined mathematically as:
    \[
        \phi = \pi \left(3 - \sqrt{5}\right)
    \]

    \item \textbf{Z-coordinate}: The vertical position \(z_i\) is evenly spaced along the hemisphere's vertical axis, defined by:
    \[
        z_i = R \left(1 - \frac{i}{N}\right)
    \]
    where \(z_i\) ranges from \(R\) (when \(i=0\)) down to \(R/N\) (when \(i=N-1\)), covering the upper hemisphere without reaching the equatorial plane.

    \item \textbf{Azimuthal Angle (\(\theta_i\))}: For each camera index \(i\), the azimuthal angle \(\theta_i\) around the vertical axis (measured from the positive \(X\)-axis in the horizontal \(XY\)-plane) is given by:
    \[
         \theta_i = (\phi\ \times i) \bmod 2\pi; \textrm{where}, ~i\, \textrm{varies from}  \{0, 1, \dots, N-1\} 
    \]
    For instance, since we index the cameras starting from 0, the fifth camera corresponds to index 5. We can compute $\theta_5$ that corresponds to the fifth camera as $\pi \left(3 - \sqrt{5}\right) \times 5  \bmod 2\pi \approx 1.81966 \pi $
    
    \item \textbf{Radius on the XY-plane}: The horizontal radial distance \(r_i\) at height \(z_i\) is:
    \[
        r_i = \sqrt{R^2 - z_i^2}
    \]

    \item \textbf{Cartesian Coordinates}: With the radial and angular coordinates established, each camera position is:
    \[
        \mathbf{p}_i = \bigl(r_i \cos\theta_i,\; r_i \sin\theta_i,\; z_i\bigr).
    \]
\end{enumerate}

The resulting set of camera positions \(\mathbf{p}_i\) are uniformly distributed across a hemisphere of radius \( R \), providing balanced spatial coverage. This mathematical formulation aligns precisely with the geometric representation provided in Figure \ref{fig:fibonacci_sphere_sampling}, clearly illustrating the relationships between \(\phi\), \( z_i \), \( r_i \), \(\theta_i\), and the resultant camera positions \(\mathbf{p}_i\).

%

\subsection{Technical Details on CMU Multiview Dataset Calibration and Coordinate Conversion}\label{Asubsec:coordinate_conv}

To make the coordinate system used in CMU Panoptic dataset \cite{joo2015panoptic} compatible with our coordinate system setting, we process each high-definition (HD) camera entry from their provided JSON dataset file by introducing the following steps:

\begin{enumerate}[leftmargin=*]

\item \textbf{Intrinsic Parameters Extraction:} 
We first extract the image resolution parameters $w$ (width), $h$ (height), and the intrinsic camera matrix $K$, defined as:
\[
K = \begin{bmatrix}
f_x & 0 & c_x \\
0 & f_y & c_y \\
0 & 0 & 1
\end{bmatrix},
\]
alongside radial distortion coefficients $(k_1, k_2, k_3)$ and tangential distortion coefficients $(p_1, p_2)$. We assign zero values when distortion parameters, i.e., $(k_1, k_2, k_3, p_1, p_2)$ are absent.

\item \textbf{Field of View Computation:}
Next, we calculate the horizontal and vertical fields of view from the intrinsic parameters as:
\[
\text{camera\_angle}_x = 2\arctan\left(\frac{w/2}{f_x}\right),\quad
\text{camera\_angle}_y = 2\arctan\left(\frac{h/2}{f_y}\right).
\]

\item \textbf{Extrinsic Matrix Formation:}
The provided rotation $R$ and translation $t$ vectors define a world-to-camera transformation matrix:
\[
M_{w\rightarrow c} = \begin{bmatrix}
R & t \\
0 & 1
\end{bmatrix}.
\]
We invert this matrix to obtain the camera-to-world transformation:
\[
M_{c\rightarrow w} = M_{w\rightarrow c}^{-1}.
\]

\item \textbf{Coordinate System Alignment:}
Since the original CMU Panoptic dataset \cite{joo2015panoptic} adopts a Y-up coordinate convention, we convert it to a standard Z-up, Y-forward coordinate system. This transformation is performed using two sequential rotations:

\[
R_{y\rightarrow z} =
\begin{bmatrix}
1 & 0 & 0 & 0 \\
0 & 0 & 1 & 0 \\
0 & -1 & 0 & 0 \\
0 & 0 & 0 & 1
\end{bmatrix},\quad
R_{x,180} =
\begin{bmatrix}
1 & 0 & 0 & 0 \\
0 & -1 & 0 & 0 \\
0 & 0 & -1 & 0 \\
0 & 0 & 0 & 1
\end{bmatrix}.
\]

The final aligned camera-to-world transformation matrix is then computed as:
\[
M'_{c\rightarrow w} = R_{x,180} \, R_{y\rightarrow z} \, M_{c\rightarrow w}.
\]

\end{enumerate}

By following the outlined steps, we generate camera configurations fully compatible with our corrdinate system. The resulting dataset entries ensure consistency in intrinsic calibration, distortion correction, image alignment, and world-coordinate system compatibility for effective training of ours as well as other approaches that uses similar coordinate system.
\subsection{Model Training Technical Details}\label{Asubsec:model_training}

\paragraph{Spatial bounding box.}
To eliminate empty-space sampling and stabilise optimisation, we constrain every
scene to the axis-aligned cube
\[
  \mathcal{B} = [-B,B]^3, \qquad B=2.
\]
All rays are clipped to their entry/exit points
\(t_{\mathrm{in/out}}\) with respect to \(\mathcal{B}\), so subsequent
integration is restricted to the interval
\([t_{\mathrm{in}},\,t_{\mathrm{out}}]\subseteq[t_{\min},t_{\max}]\).

\paragraph{Volume rendering.}
For a camera ray \(\mathbf r(t)=\mathbf o+t\mathbf d\),
we follow the standard formulation
\[
  C(\mathbf r)=
  \int_{t_{\mathrm{in}}}^{t_{\mathrm{out}}}
  T^{t}(t)\sigma(\mathbf r(t))c^{t}(\mathbf r(t))dt \quad
  T^{t}(t)=\exp\Bigl(-\int_{t_{\mathrm{in}}}^{t}\sigma^{t}(\mathbf r(s)) ds\Bigr),
\]
which we approximate with a Riemann sum of step size \(\delta\):
\[
  C(\mathbf r)\approx
  \sum_{i=1}^{N_{\mathcal B}}
  T_i\sigma_ic_i\delta,
  \qquad
  N_{\mathcal B} = \bigl\lceil (t_{\mathrm{out}}-t_{\mathrm{in}})/\delta \bigr\rceil.
\]

\paragraph{Camera scaling.}
Because the original camera centres \(\mathbf p\) may fall outside the
bounding box \(\mathcal{B}\), we apply a global scale factor
\texttt{scale} such that the rescaled positions
\[
  \mathbf p' = \texttt{scale} ~\mathbf {p}
\]
satisfy \(\mathbf{p}' \in \mathcal{B}\).

\paragraph{Network architecture.}
Our model comprises three sequential modules that operate on \emph{learned
positional embeddings} and an \emph{encoded view direction}:

\begin{itemize}
     \item \textbf{Positional encoder.}
          For each sample point
          \(\mathbf{x}=(x,y,z)^\mathsf{T}\in\mathcal B\),
          we first normalise its coordinates to \([0,1]^3\) and then feed
          them into a multi-resolution hash grid with
          \(L=16\) levels and \(C=2\) channels per entry.
          Let \(R_{\min}=16\) and \(R_{\max}=2048 \times B\) be the grid
          resolutions at levels \(0\) and \(L\!-\!1\), respectively.
          The geometric scale factor is
          \[
            \alpha = (R_{\max}/R_{\min})^{1/(L-1)}, \qquad
            R_l = R_{\min}\,\alpha^l.
          \]
          The resulting positional embedding has dimension
          \(L\times C = 32\).

  \item \textbf{Density MLP.}
        A two-layer network
        \[
          \mathrm{Linear}(32,64)\!\xrightarrow{\text{ReLU}}\!
          \mathrm{Linear}(64,1{+}15)
        \]
        yields the volume density \(\sigma\) and a 15-D geometric feature
        \(\mathbf f\).

  \item \textbf{Color MLP.}
        The view direction is expanded to a
        \(D_{\mathrm{dir}}{=}16\)-D spherical-harmonics vector,
        concatenated with \(\mathbf f\), and processed by
        \[
          \mathrm{Linear}(D_{\mathrm{dir}}{+}15,64)\!\xrightarrow{\text{ReLU}}\!
          \mathrm{Linear}(64,64)\!\xrightarrow{\text{ReLU}}\!
          \mathrm{Linear}(64,3)\!\xrightarrow{\text{Sigmoid}}
        \]
        to predict RGB.
\end{itemize}

\paragraph{Parameter count.}
Each model has a single hash table of \(2^{22}=4{,}194{,}304\) entries
(\(2\) channels each) and, together with the two MLPs, amounts to approximately 12.7 million learnable parameters.

\paragraph{Training Configurations for Each Dataset .} Detailed training configurations for each dataset are:

\begin{itemize}[noitemsep,topsep=1pt]
  \item \textbf{Dancing-Walking-Standing Dataset:} Each MLP model was trained for \(19,000\) iterations, with \texttt{scale = 0.3}.
  \item \textbf{Soccer Penalty Kick Dataset:} Each MLP model was trained for \(16,500\) iterations, with \texttt{scale = 0.1}.
  \item \textbf{Soccer Multiplayer Dataset:} Each MLP model was trained for \(16,500\) iterations, with \texttt{scale = 0.1}.
  \item \textbf{Baseball Bat from CMU Panoptic Dataset:} Each MLP model was trained for \(14,500\) iterations, with \texttt{scale = 0.006}.
  \item \textbf{Hand Gesture from CMU Panoptic Dataset:} Each MLP model was trained for \(14,500\) iterations, with \texttt{scale = 0.006}.
\end{itemize}

\end{document}